\documentclass[10pt,twocolumn,letterpaper]{article}

\usepackage[pagenumbers]{cvpr} % To force page numbers, e.g. for an arXiv version

\usepackage[dvipsnames]{xcolor}

\definecolor{cvprblue}{rgb}{0.21,0.49,0.74}
\usepackage[pagebackref,breaklinks,colorlinks,citecolor=cvprblue]{hyperref}

\usepackage{booktabs}
\usepackage{nicematrix}
\usepackage{algorithm}
\usepackage{algorithmic}
\usepackage[frozencache,cachedir=mintedcache]{minted}
\setminted{breaklines}
\usepackage{multirow}

\usepackage{setspace}
\usepackage[accsupp]{axessibility}
\def\paperID{7545} % *** Enter the Paper ID here
\def\confName{CVPR}
\def\confYear{2024}

\title{Improving Plasticity in Online Continual Learning via Collaborative Learning}

\author{Maorong Wang$^{1}$ \quad Nicolas Michel$^{2}$ \quad Ling Xiao$^{1}$ \quad Toshihiko Yamasaki$^{1}$ \vspace{0.3em} \\
{\normalsize $^1$The University of Tokyo} \quad
{\normalsize $^2$Univ Gustave Eiffel, CNRS, LIGM} \quad \\
{\normalsize \{ma\_wang, ling, yamasaki\}@cvm.t.u-tokyo.ac.jp, nicolas.michel@univ-eiffel.fr} \quad \\
}

\begin{document}
\maketitle
\begin{abstract}
Online Continual Learning (CL) solves the problem of learning the ever-emerging new classification tasks from a continuous data stream. Unlike its offline counterpart, in online CL, the training data can only be seen once. Most existing online CL research regards catastrophic forgetting (i.e., model stability) as almost the only challenge. In this paper, we argue that the model's capability to acquire new knowledge (i.e., model plasticity) is another challenge in online CL. While replay-based strategies have been shown to be effective in alleviating catastrophic forgetting, there is a notable gap in research attention toward improving model plasticity. To this end, we propose Collaborative Continual Learning (CCL), a collaborative learning based strategy to improve the model's capability in acquiring new concepts. Additionally, we introduce Distillation Chain (DC), a collaborative learning scheme to boost the training of the models. We adapt CCL-DC to existing representative online CL works. Extensive experiments demonstrate that even if the learners are well-trained with state-of-the-art online CL methods, our strategy can still improve model plasticity dramatically, and thereby improve the overall performance by a large margin. 
% The source code is included in the supplementary material and will be publicly available upon acceptance.
% For example, CCL-DC can improve the accuracy of the state-of-the-art method (OCM) on Tiny-ImageNet (M=5000) by 24\% from 26.74\% to 33.17\%. 
The source code of our work is available at \href{https://github.com/maorong-wang/CCL-DC}{https://github.com/maorong-wang/CCL-DC}.
\end{abstract}
  
\section{Introduction}
\label{sec:intro}

% In the past few years, we have witnessed the success of deep neural networks in various computer vision tasks. However, current model optimization techniques rely heavily on two important priors: (1) 

Continual Learning (CL)~\cite{chen2018lifelong, parisi2019continual, wang2023comprehensive, de2021continual} aims to incrementally learn a sequence of tasks and enhance the neural network's performance on the current tasks, without forgetting previously learned knowledge. CL can be done in two different manners~\cite{wang2023comprehensive,aljundi2019gradient}: \textit{offline} and \textit{online}. In offline CL, the learner can have infinite access to all the training data of the current task that it trains on and may go through the data for any epoch. Contrary to offline CL, in online CL, the training data for each task also comes continually in a data stream, and the learner can only see the training data once. Apart from the learning approach, there are also three different CL scenarios~\cite{hsu2018re, van2019three, masana2022class}: Task-Incremental Learning (TIL), Domain-Incremental Learning (DIL), and Class-Incremental learning (CIL). In this paper, we focus on the CIL setting in online CL.

\begin{figure}[t]
  \centering
   \includegraphics[width=0.9\linewidth]{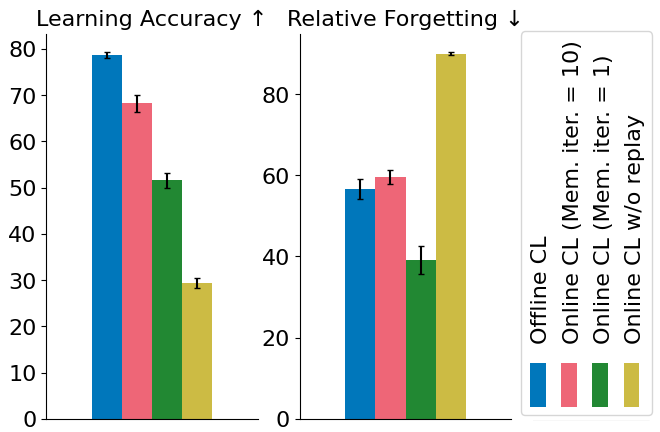}
   \vspace{-7pt}
   \caption{The comparison of plasticity (learning accuracy) and stability (relative forgetting, our metric proposed in Sec.~\ref{sec:tradeoff}) of Experience Replay (ER)~\cite{rolnick2019experience} under varisous settings on CIFAR-100. For experiments with memory replay, the size of the memory buffer is set to 2,000. We can witness a plasticity gap between offline CL and online CL, even with memory replay and multiple update trick (memory iteration $>$ 1).}
   \vspace{-4pt}
   \label{fig:plasticity_gap}
\end{figure}

Various online CL methods~\cite{rolnick2019experience, caccia2021new, buzzega2020dark, guo2022online, guo2023dealing, wei2023online, michel2023learning} have been proposed to help the models learn continually on one-epoch data stream, with alleviated forgetting. Among them, replay-based methods have shown remarkable success, and current state-of-the-art methods rely heavily on memory replay to mitigate catastrophic forgetting~\cite{mccloskey1989catastrophic, goodfellow2013empirical}. However, while most existing online CL research primarily only focuses on improving model stability (\textit{i.e.}, alleviating catastrophic forgetting) to achieve better overall accuracy, the importance of model plasticity (\textit{i.e.}, the capability to acquire new knowledge) is often overlooked. While in offline CL, it is possible to gain high plasticity by iterating several epochs on the current task before proceeding to the subsequent task, acquiring plasticity in online CL tends to be more challenging because the training data is only available once. As shown in Fig.~\ref{fig:plasticity_gap}, compared to learning without memory replay, the replay-based methods implicitly alleviate the low plasticity issue to some extent. Also, plasticity can be improved by employing a technique involving multiple updates on incoming samples~\cite{NIPS2019_9357}. However, the combination of memory replay and multiple updates fails to bridge the existing plasticity gap between online and offline CL, and the use of multiple updates can lead to increased catastrophic forgetting. Overall, the plasticity gap adversely affects the performance of online CL methods.

In this paper, we argue that plasticity is particularly crucial and challenging to acquire in online CL, and even more so than in offline CL. Thus, we shed light on how model plasticity and stability will impact the overall performance. Furthermore, we propose a quantitative link between plasticity, stability, and final accuracy, showing that the plasticity gap between offline and online must be reduced to improve overall performance.

Guided by the quantitative relationship, we focus ourselves on the previously overlooked plasticity perspective. Inspired by the ability of collaborative learning to accelerate the convergence in non-continual scenarios~\cite{anil2018large}, we have incorporated collaborative learning in online CL and observed a similar phenomenon. Consequently, we propose the Collaborative Continual Learning with Distillation Chain (CCL-DC), a collaborative learning scheme that can be adapted to existing online CL methods. CCL-DC consists of two key components: Collaborative Continual Learning (CCL) and Distillation Chain (DC).

CCL involves two peer continual learners to learn from the data stream simultaneously in a peer teaching manner, enhancing optimization parallelism and offering greater flexibility to the continual learners. To the best of our knowledge, CCL is the first method to incorporate collaborative learning techniques in online CL research. Moreover, to fully exploit the potential of collaborative learning in online CL scenarios, we propose DC, an entropy regularization based optimization strategy, specifically designed for online CL.  

The main contribution of this paper can be summarized as follows.

\begin{enumerate}

    \item We establish a quantitative link between plasticity, stability, and final performance. Based on this, we demonstrate that plasticity is an important obstacle in online CL, which was greatly overlooked in the previous research;
    
    \item To overcome the plasticity issue, we introduce CCL-DC, a collaborative learning based strategy that can be seamlessly integrated into the existing methods, significantly improving their performance by boosting plasticity;
    
    \item Comprehensive experiments show that CCL-DC enhances the performance of existing methods by a large margin.
\end{enumerate}

\begin{figure*}[t]
  \centering
   \includegraphics[width=0.85\linewidth]{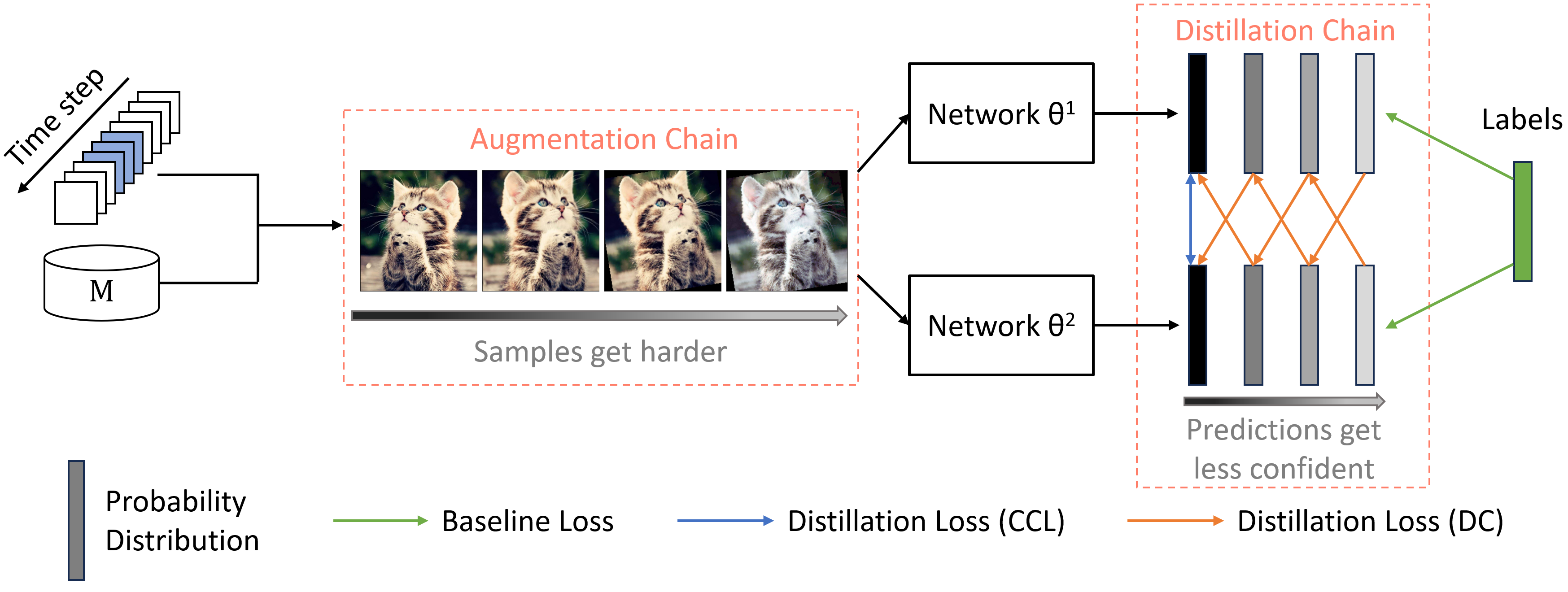}
   \caption{Overview of the proposed CCL-DC framework applied to a baseline online CL method. The proposed CCL-DC framework has two main components. The first one is CCL, which involves two peer continual learners that simultaneously learn from the data stream in a peer teaching manner. The second component, DC, generates a chain of samples with varying levels of difficulty and feeds them into models to produce a chain of logit distribution of different confidence levels. Then, in a collaborative learning approach, DC conducts distillation from \textit{less confident} predictions to \textit{more confident} predictions, to serve as a form of learned entropy regularization.}
   \label{fig:teaser}
   \vspace{-8pt}
\end{figure*}

\section{Related Work}
\label{sec:related}

\paragraph{Continual Learning.}
Continual learning can be classified into three different categories: regularization-based, parameter-isolation-based, and replay-based. Regularization-based methods~\cite{chaudhry2018riemannian, lee2017overcoming, aljundi2018memory, kirkpatrick2017overcoming, zenke2017continual} add extra regularization terms to balance the old and new tasks. Parameter-isolation-based methods~\cite{fernando2017pathnet, rusu2016progressive, serra2018overcoming, rosenfeld2018incremental, aljundi2017expert} solve the problem explicitly by dynamically allocating task-specific parameters. Replay-based methods~\cite{rolnick2019experience, caccia2021new, buzzega2020dark, guo2022online, guo2023dealing, wei2023online, michel2023learning, chaudhry2019tiny, de2019episodic} maintain a small memory buffer that stores a few old training samples. 

Among these methods, replay-based strategies have gained huge success due to their impressive performance and simplicity. ER~\cite{rolnick2019experience} is the fundamental replay-based method that leverages Cross-Entropy loss for classification and a random replay buffer. DER++~\cite{buzzega2020dark} stores the logits in the memory buffer and extends ER with the distillation of old stored logits. ER-ACE~\cite{caccia2021new} extends ER with Asymmetric Cross-Entropy loss for classification to suppress the drift of old class representations. OCM~\cite{guo2022online} leverages a replay-based strategy by maximizing the mutual information between old and new class representations. GSA~\cite{guo2023dealing} solves cross-task class discrimination with replay-based strategy and Gradient Self Adaption. OnPro~\cite{wei2023online} uses online prototype learning to address shortcut learning and alleviate catastrophic forgetting.

These replay-based methods propose different strategies for alleviating catastrophic forgetting and improving the model stability. However, the importance of the model plasticity is greatly neglected in their research, despite their success in terms of final performance. In our work, these methods serve as the baselines and we adapted our strategy to these baselines to show the effectiveness of our proposed approach.

%In our work, we focus on the perspective of the model's plasticity and propose CCL-DC, a strategy that can boost the model's ability to learn new concepts. Even if the continual learners are trained with state-of-the-art online CL methods, 

\vspace{-8pt}
\paragraph{Collaborative Learning.}
Collaborative learning~\cite{zhang2018deep, guo2020online, anil2018large, zhu2018knowledge, song2018collaborative} originates from online knowledge distillation (KD). Different from the conventional KD methods, online KD involves training a cohort of deep networks from scratch in a peer-teaching manner. During the training process, the models imitate their peers and guides the training of other models simultaneously. Deep Mutual Learning (DML)~\cite{zhang2018deep} suggests peer student models learn from each other through the logit distillation between the probability distributions. Codistillation~\cite{anil2018large} is similar to DML and suggests the ensemble of peer networks can further improve the performance. More importantly, Codistillation shows that online KD can help the model converge faster on non-continual scenarios. 

Despite the success of collaborative learning in non-continual scenarios, due to the lack of focus on plasticity, the research on collaborative learning in CL is still limited. To the best of our knowledge, there is no existing research using the collaborative learning technique to boost the training of online CL. Moreover, in our work, we introduce the DC, an entropy regularization based optimization strategy, specifically tailored to fully leverage the potential of collaborative learning in online CL scenarios. 

\section{Plasticity and Stability in online CL}
\label{sec:tradeoff}
In this section, we revise the metric for model plasticity and propose a novel metric for model stability. In addition, we quantitatively analyze the impact of model plasticity and stability on the final performance.

\subsection{Model Plasticity}
The model plasticity quantifies the learner's ability to learn new knowledge when a new task arrives. Several different metrics have been proposed to measure the model plasticity~\cite{wang2023comprehensive, chaudhry2018riemannian, lopez2017gradient, riemer2018learning}. In our work, we evaluate the model plasticity with Learning Accuracy (LA)~\cite{riemer2018learning}. Formally, the LA for the $j$-th task is defined as:
\begin{equation}
\begin{split}
l_j = a_j^j, 
\end{split}
\label{eq:LA}
\end{equation}
where $a_j^i$ is the accuracy evaluated on the test set of task $j$ after training the network from task 1 to task $i$. For an overall metric normalized against all tasks, the averaged LA is written as $LA = \frac{1}{T} \sum_{j=1}^T l_j$, and $T$ is the number of tasks in total.

\subsection{Model Stability}
The stability assesses the extent of knowledge retention or loss within a model in its present state. The most commonly used metric in previous CL research is the Forgetting Measure (FM)~\cite{chaudhry2018riemannian}. Intuitively, FM for the $j$-th task, denoted as $fm_j^k$, quantifies the decline in performance  on a given task $j$, after the model is trained on task $k$, relative to its highest past performance on task $j$:
\begin{equation}
fm_j^k = \max_{i\in\{1,...,k-1\}} (a_j^i-a_j^k),~\forall j < k. 
\label{eq:FM}
\end{equation}

For the overall metric obtained across all tasks, FM can be expressed as:
\begin{equation}
FM = \frac{1}{T-1} \sum_{j=1}^{T-1} fm_j^T.
\label{eq:FM_all}
\end{equation}

In our work, we propose a novel stability metric, termed Relative Forgetting (RF), as an alternative to the traditional Forgetting Measure (FM). RF measures how much \textit{proportion} of performance the model forgets. Specifically, RF for the $j$-th task after training on task $k$ is defined as:
\begin{equation}
f_j^k = \max_{i\in\{1,...,k\}} \left(1-\frac{a_j^k}{a_j^i} \right),~\forall j \le k.
\label{eq:RF}
\end{equation}

The overall metric averaged across all tasks can be written as:
\begin{equation}
RF = \frac{1}{T} \sum_{j=1}^{T} f_j^T.
\label{eq:RF_all}
\end{equation}

\begin{table}[t]
    \centering
    \resizebox{\linewidth}{!}{
        \begin{tabular}{c | c c c c c | c c}
            $a_j^i$      &$\mathcal{T}_1$  & $\mathcal{T}_2$ & $\mathcal{T}_3$ & $\mathcal{T}_4$ &$\mathcal{T}_5$ & $FM_k$ & $RF_k$ \\ [2pt]
            \hline
            $\mathcal{T}_1$ & $\textcolor{teal}{30} / \textcolor{brown}{15}$ & - & - & - & - & - & - \\ [2pt]
            $\mathcal{T}_2$ & $\textcolor{teal}{25} / \textcolor{brown}{12.5}$ & $\textcolor{teal}{25} / \textcolor{brown}{12.5}$ & - & - & - & $\textcolor{teal}{5} / \textcolor{brown}{2.5}$ & $\textcolor{teal}{8.33} / \textcolor{brown}{8.33}$ \\ [2pt]
            $\mathcal{T}_3$ & $\textcolor{teal}{20} / \textcolor{brown}{10}$ & $\textcolor{teal}{20} / \textcolor{brown}{10}$ & $\textcolor{teal}{20} / \textcolor{brown}{10}$ & - & - & $\textcolor{teal}{7.5} / \textcolor{brown}{3.75}$ & $\textcolor{teal}{17.78} / \textcolor{brown}{17.78}$ \\ [2pt]
            $\mathcal{T}_4$ & $\textcolor{teal}{15} / \textcolor{brown}{7.5}$ & $\textcolor{teal}{15} / \textcolor{brown}{7.5}$ & $\textcolor{teal}{15} / \textcolor{brown}{7.5}$ & $\textcolor{teal}{15} / \textcolor{brown}{7.5}$ & - & $\textcolor{teal}{10} / \textcolor{brown}{5}$ & $\textcolor{teal}{28.75} / \textcolor{brown}{28.75}$ \\ [2pt]
            $\mathcal{T}_5$ & $\textcolor{teal}{10} / \textcolor{brown}{5}$ & $\textcolor{teal}{10} / \textcolor{brown}{5}$ & $\textcolor{teal}{10} / \textcolor{brown}{5}$ & $\textcolor{teal}{10} / \textcolor{brown}{5}$ & $\textcolor{teal}{10} / \textcolor{brown}{5}$ & $\textcolor{teal}{12.5} / \textcolor{brown}{6.25}$ & $\textcolor{teal}{42} / \textcolor{brown}{42}$
        \end{tabular}
    }
    \vspace{-.8em}
    \caption{Forgetting Measure (\%) and Relative Forgetting (\%) of two example learners (Learner 1 in \textcolor{teal}{teal} and Learner 2 in \textcolor{brown}{brown}) in a continual setting of five tasks with two classes per task. Each element $a_j^i$ at row $j$ and column $i$ is the per task accuracy evaluated on the test set of task $j$ after training the network from task 1 to task $i$.}
    \vspace{-1.4em}
    \label{tab:example}
\end{table}

There are two advantages for shifting from absolute forgetting to relative forgetting: 

\begin{enumerate}
    \item RF is fairer for methods with higher plasticity. 
    As shown in Table~\ref{tab:example}, when comparing two continual learners, one with high plasticity and the other with low, using both forgetting metrics, it is observed that despite having identical RF values, the model with lower plasticity shows a lower FM. This indicates that the FM metric might unfairly favor models with poorer initial performance, even when the relative decline in performance is equivalent across models; 
    \item RF facilitates a quantitative analysis of the relationship between model stability and final performance. This enables a more nuanced understanding of how stability influences overall model efficacy, thereby guiding improvements in continual learning strategies. 
\end{enumerate}

\subsection{Impact on the Overall Performance}

For online CL, the model's final average accuracy (AA) is the most vital metric. In this subsection, we try to show how the model plasticity and stability will impact the final performance quantitatively. 

The model's final average accuracy can be calculated as:
\begin{equation}
AA = \frac{1}{T} \sum_{j=1}^{T} a_j^T.
\label{eq:AA}
\end{equation}
With the establishment of our plasticity metric (LA) and stability metric (RF), we can easily find the relationship between learning accuracy, relative forgetting, and accuracy:
\begin{equation}
a_j^i \ge l_j \times (1-f_j^i),
\label{eq:AA_classwise}
\end{equation}
assuming equality when $a_j^j = \max_{i\in\{1,...,j\}} a_j^i$. When extending the class-specific final accuracy $a_j^T$ to the final AA, we consider the dot product of the LA vector $[l_1,...,l_T]$ with the RF vector $[f_1^T,...,f_T^T]$, which simplifies the calculation. More practically, this relationship can be approximated with:

\begin{equation}
AA \gtrapprox LA \times (1-RF).
\label{eq:AA_all}
\end{equation}

As indicated by Eq.~\ref{eq:AA_all}, the lower bound of the final performance is proportional to $LA$ and $1-RF$, which suggests that both plasticity (LA) and stability (RF) play a crucial role in the final accuracy. Our findings reveal the importance of the model plasticity which was neglected in the past. And it can serve as a good guide for future online CL research.

\section{Proposed Method}
In this section, we first justify our motivations. Then, we introduce our proposed strategy: Collaborative Continual Learning and Distillation Chain. Finally, we show how to adapt our proposed strategy to the existing online CL methods and boost their plasticity. 
\label{sec:method}

\subsection{Motivation Justification}
\begin{figure}
  \centering
   \includegraphics[width=0.87\linewidth]{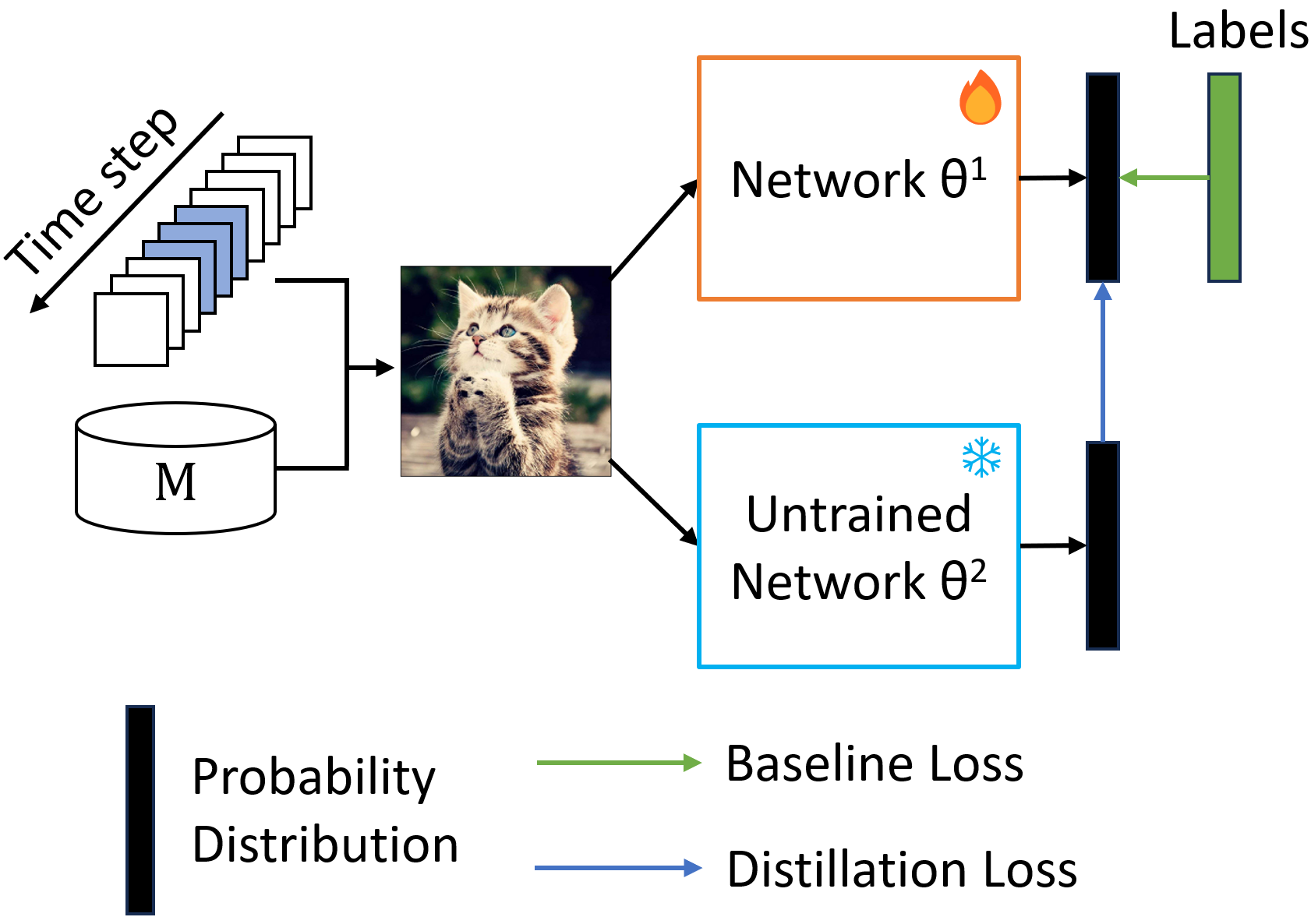}
   \vspace{-1.1em}
   \caption{Conceptual diagram of the training framework, when distilling from an \textit{untrained} network $\theta^2$ to suppress the confidence of network $\theta^1$.}
   \vspace{-1em}
   \label{fig:rand}
\end{figure}

\begin{table}
    \centering
    \resizebox{0.9\linewidth}{!}{
    \begin{tabular}{lccc}
\toprule
\multicolumn{1}{c}{Dataset} & \multicolumn{3}{c}{CIFAR-100} \\
\cmidrule(lr){1-1}
\cmidrule(lr){2-4}
\multicolumn{1}{c}{Memory Size $M$} & M=1000 & M=2000 & M=5000 \\
\midrule
ER & 24.47{\scriptsize ±0.72} & 31.89{\scriptsize ±1.45} & 39.41{\scriptsize ±1.81} \\
ER + Untrained Distillation & 27.07{\scriptsize ±1.20} & 34.84{\scriptsize ±0.64} & 41.15{\scriptsize ±1.16}\\
\bottomrule
\end{tabular}
    }
    \vspace{-.8em}
    \caption{The performance of network $\theta^1$ when distilling from untrained network $\theta^2$ on CIFAR-100. All numbers are average over 10 runs.}
    \vspace{-1.4em}
    \label{tab:rand}
\end{table}

\paragraph{Plasticity matters in Online CL.}
Online continual learners are designed to continuously adapt to non-stationary data streams, efficiently acquiring new knowledge while retaining previously learned information. In current online CL research, the emphasis has predominantly been on mitigating catastrophic forgetting. Even among state-of-the-art methods such as~\cite{guo2022online, wei2023online}, plasticity is often compromised in the quest for improved stability. However, our finding in Sec.~\ref{sec:tradeoff} shows that both plasticity and stability are crucial for achieving satisfactory final performance, with plasticity proving particularly difficult to achieve in online CL settings. To this end, we deliberately prioritize the plasticity aspect. 

The potential of collaborative learning to improve convergence in non-continual scenarios~\cite{anil2018large} positions it as a promising candidate for enhancing plasticity. With the apparent lack of focus on plasticity, collaborative learning has yet to be leveraged to boost convergence of online continual learners. In our research, we propose to exploit collaborative learning convergence properties for improving plasticity. We find that similar to non-continual scenarios, collaborative learning strategy can boost convergence by allowing more parallelism in the training and more maneuverability of the continual learners.

\vspace{-10pt}
\paragraph{Overconfidence hurts Online CL.} In conventional supervised learning, it is well established that excessive confidence can harm the generalization ability. To tackle this issue, many methods such as label smoothing~\cite{szegedy2016rethinking}, knowledge distillation~\cite{yuan2020revisiting}, and confidence penalty~\cite{pereyra2017regularizing} are proposed. 

In online CL, we find a counter-intuitive, yet similar phenomenon. As shown in Fig.~\ref{fig:rand}, while we train network $\theta^1$ with the classification loss, we initiate another network $\theta^2$ at the beginning of the training. During the continual training, we stagnate the network $\theta^2$ (\textit{i.e.}, Kaiming initialize and stay untrained) and train the network $\theta^1$ with both classification loss and the distillation loss (Kullback-Leibler divergence loss) from untrained network $\theta^2$. The experimental result in Table~\ref{tab:rand} shows a decent performance gain compared with independent training. The distillation from the prediction of \textit{untrained} network $\theta^2$ to network $\theta^1$ serves as a regularization to suppress the overall confidence, and the performance gain under this condition makes it evident that, in continual scenarios, overconfidence will also harm the performance.

To tackle the overconfidence problem, we propose Distillation Chain (DC), an entropy regularizer to suppress the overall confidence level.

\subsection{Collaborative Continual Learning}

The introduced Collaborative Continual Learning (CCL) enables more parallelism and flexibility in training online continual learners, and it is the key to improving the model plasticity and the final performance. As shown in Fig.~\ref{fig:teaser}, CCL involves two peer continual learners of the same architecture and optimizer setting training in a peer-teaching manner. In the training phase, networks are supervised with both the ground truth label and the predictions of their peers. In the inference phase, models can either make predictions collaboratively with ensemble methods~\cite{anil2018large} to get a better performance or predict independently for the sake of computation efficiency. If we denote two networks in CCL as $\theta^1$ and $\theta^2$, we formulate our loss to network $\theta^1$ as:
\begin{equation}
\begin{aligned}
\mathcal{L}_{CCL}^1 =&\lambda_1 \cdot \mathcal{L}_{cls}(\theta^1(X), y) \\
    +& \lambda_2 \cdot D_{KL}(\theta^1(X)/\tau, \theta^2(X)/\tau),
\end{aligned}
\label{eq:CCL}
\end{equation}
where $(X,y)$ is the data-label pair, $\mathcal{L}_{cls}(\cdot)$ is the classification loss in the baseline method CCL adapts to, $D_{KL}(\cdot)$ is the Kullback-Leibler divergence, $\lambda_1$ and $\lambda_2$ are the balancing hyperparameters and $\tau$ is the temperature hyperparameter. Note that the network $\theta^2$ should be trained with $\mathcal{L}_{CCL}^2$, respectively.

\subsection{Distillation Chain}

To fully take advantage of CCL, we propose Distillation Chain (DC), an entropy regularizaion based strategy explicitly designed for online CL. As illustrated in Fig.~\ref{fig:teaser}, DC comprises two steps: (1) generating a chain of samples with different levels of difficulty~\cite{soviany2022curriculum} using data augmentation, and (2) distillation of logit distribution from \textit{harder} samples to \textit{easier} samples in a collaborative learning way. 

%The main motivation of DC originates from the idea of entropy regularization-based optimization strategies, like label smoothing~\cite{szegedy2016rethinking}, knowledge distillation~\cite{yuan2020revisiting}, and confidence penalty~\cite{pereyra2017regularizing}, where we find that overconfidence will hurt performance in non-continual scenarios. 
As shown in Table~\ref{tab:rand}, we observed that overconfidence will hurt the performance in continual training. To tackle the problem, DC uses data augmentation strategies to generate samples with different levels of difficulty and produces logit distribution with different confidence. The distillation from \textit{less confident} predictions to \textit{more confident} predictions weakens the overall confidence of the network and benefits the performance by improving the generalization capability.

In our work, we use a geometric distortion comprised of RandomCrop and RandomHorizontalFlip as the first step of DC augmentation. After that, we use RandAugment~\cite{cubuk2020randaugment} for the subsequent augmentations and we involve two hyperparameters $N$ and $M$ for RandAugment. We take three augmentation steps and distill the logit distribution from the teacher with \textit{harder} samples to the student with \textit{easier} samples. We formulate our loss with DC to network $\theta^1$ as:
\begin{equation}
\begin{aligned}
\mathcal{L}_{DC}^1 = &\lambda_1 \sum_{i=1}^3 \mathcal{L}_{cls}(\theta^1(X_i), y)\\
    +&\lambda_2 \sum_{i=1}^3 D_{KL}(\theta^1(X_{i-1})/\tau, \theta^2(X_i)/\tau),
\end{aligned}
\label{eq:DC}
\end{equation}
where $X_i$ is the augmentation of input sample $X$ after $i$ augmentation steps.

\begin{spacing}{0.8}
\begin{algorithm}[t]
\scriptsize
\begin{minted}{python}
# model1: student model
# model2: teacher model
# optim1: optimizer for student model
# cls: classification loss in baseline 
for x, y in dataloader:
  # Baseline loss
  loss_baseline = criterion_baseline(model1, x, y)
  
  # DC Augmentation
  x1 = geometric_distortion(x) 
  x2 = RandAugment(x1, N, M)
  x3 = RandAugment(x2, N, M)  

  # CCL-DC loss
  ls, ls1, ls2, ls3 = model1(x, x1, x2, x3)
  lt, lt1, lt2, lt3 = model2(x, x1, x2, x3) # no grad

  loss_cls = cls(ls, y) + cls(ls1, y) + cls(ls2, y) + cls(ls3, y)
  loss_ccl = kl_div(ls/t, lt/t) # temperature t
  loss_dc = kl_div(ls/t, lt1/t) + kl_div(ls1/t, lt2/t) + kl_div(ls2/t, lt3/t)

  loss_ours = lam1*loss_cls + lam2*(loss_ccl + loss_dc)
  loss = loss_baseline + loss_ours
  
  optim1.zero_grad()
  loss.backward()
  optim1.step()
\end{minted}
\vspace{-4pt}
\caption{PyTorch-like pseudo-code of CCL-DC to integrate to other baselines.}
\label{code:pseudo_code}
\end{algorithm}
\end{spacing}

\begin{table*}
    \centering
    \resizebox{\textwidth}{!}{
    \begin{tabular}{lccccccccc}
\toprule
\multicolumn{1}{c}{Dataset} & \multicolumn{2}{c}{CIFAR10} & \multicolumn{3}{c}{CIFAR100} & \multicolumn{3}{c}{Tiny-ImageNet} & ImageNet-100\\
\cmidrule(lr){1-1}
\cmidrule(lr){2-3}
\cmidrule(lr){4-6}
\cmidrule(lr){7-9}
\cmidrule(lr){10-10}
\multicolumn{1}{c}{Memory Size $M$} & 500 & 1000 & 1000 & 2000 & 5000 & 2000 & 5000 & 10000 & 5000\\
\midrule
ER~\cite{rolnick2019experience} & 56.68{\scriptsize ±1.89} & 62.32{\scriptsize ±4.13} & 24.47{\scriptsize ±0.72} & 31.89{\scriptsize ±1.45} & 39.41{\scriptsize ±1.81} & 10.82{\scriptsize ±0.79} & 19.16{\scriptsize ±1.42} & 24.71{\scriptsize ±2.52} & 33.30{\scriptsize ±1.74}\\
ER + Ours & \textbf{66.43{\scriptsize ±2.48}} & \textbf{74.10{\scriptsize ±1.71}} & \textbf{33.43{\scriptsize ±1.06}} & \textbf{44.45{\scriptsize ±1.04}} & \textbf{53.81{\scriptsize ±1.16}} & \textbf{16.56{\scriptsize ±1.63}} & \textbf{29.39{\scriptsize ±1.23}} & \textbf{37.73{\scriptsize ±0.85}} & \textbf{43.11{\scriptsize ±1.49}}\\
\midrule
DER++~\cite{buzzega2020dark} & 58.04{\scriptsize ±2.30} & 64.02{\scriptsize ±1.92} & 25.09{\scriptsize ±1.41} & 32.33{\scriptsize ±2.66} & 38.31{\scriptsize ±2.28} & 8.73{\scriptsize ±1.58} & 17.95{\scriptsize ±2.49} & 19.40{\scriptsize ±3.71} & 34.75{\scriptsize ±2.23}\\
DER++ + Ours & \textbf{68.79{\scriptsize ±1.42}} & \textbf{74.25{\scriptsize ±1.10}} & \textbf{34.36{\scriptsize ±0.89}} & \textbf{43.52{\scriptsize ±1.35}} & \textbf{52.95{\scriptsize ±0.86}} & \textbf{10.99{\scriptsize ±1.39}} & \textbf{21.68{\scriptsize ±1.94}} & \textbf{28.01{\scriptsize ±2.46}} & \textbf{45.70{\scriptsize ±1.32}}\\
\midrule
ER-ACE~\cite{caccia2021new}  & 53.26{\scriptsize ±3.04} & 59.94{\scriptsize ±2.40} & 28.36{\scriptsize ±1.99} & 34.21{\scriptsize ±1.53} & 39.39{\scriptsize ±1.31} & 13.56{\scriptsize ±1.00} & 20.84{\scriptsize ±0.43} & 25.92{\scriptsize ±1.07} & 38.37{\scriptsize ±1.20}\\
ER-ACE + Ours & \textbf{70.08{\scriptsize ±1.38}} & \textbf{75.56{\scriptsize ±1.14}} & \textbf{37.20{\scriptsize ±1.15}} & \textbf{45.14{\scriptsize ±1.00}} & \textbf{53.92{\scriptsize ±0.48}} & \textbf{18.32{\scriptsize ±1.49}} & \textbf{26.22{\scriptsize ±2.01}} & \textbf{32.23{\scriptsize ±1.70}} & \textbf{45.15{\scriptsize ±1.94}}\\
\midrule
OCM~\cite{guo2022online} & 68.19{\scriptsize ±1.75} & 73.15{\scriptsize ±1.05} & 28.02{\scriptsize ±0.74} & 35.69{\scriptsize ±1.36} & 42.22{\scriptsize ±1.06} & 18.36{\scriptsize ±0.95} & 26.74{\scriptsize ±1.02} & 31.94{\scriptsize ±1.19} & 23.67{\scriptsize ±2.36}\\
OCM + Ours & \textbf{74.14{\scriptsize ±0.85}} & \textbf{77.66{\scriptsize ±1.46}} & \textbf{35.00{\scriptsize ±1.15}} & \textbf{43.34{\scriptsize ±1.51}} & \textbf{51.43{\scriptsize ±1.37}} & \textbf{23.36{\scriptsize ±1.18}} & \textbf{33.17{\scriptsize ±0.97}} & \textbf{39.25{\scriptsize ±0.88}} & \textbf{43.19{\scriptsize ±0.98}}\\
\midrule
GSA~\cite{guo2023dealing} & 60.34{\scriptsize ±1.97} & 66.54{\scriptsize ±2.28} & 27.72{\scriptsize ±1.57} & 35.08{\scriptsize ±1.37} & 41.41{\scriptsize ±1.65} & 12.44{\scriptsize ±1.17} & 19.59{\scriptsize ±1.30} & 25.34{\scriptsize ±1.43} & 41.03{\scriptsize ±0.99}\\
GSA + Ours & \textbf{68.91{\scriptsize ±1.68}} & \textbf{75.78{\scriptsize ±1.16}} & \textbf{35.56{\scriptsize ±1.39}} & \textbf{44.74{\scriptsize ±1.32}} & \textbf{55.39{\scriptsize ±1.09}} & \textbf{16.70{\scriptsize ±1.66}} & \textbf{28.11{\scriptsize ±1.70}} & \textbf{37.13{\scriptsize ±1.75}} & \textbf{44.28{\scriptsize ±1.16}}\\
\midrule
OnPro~\cite{wei2023online} & 70.47{\scriptsize ±2.12} & 74.70{\scriptsize ±1.51} & 27.22{\scriptsize ±0.77} & 33.33{\scriptsize ±0.93} & 41.59{\scriptsize ±1.38} & 14.32{\scriptsize ±1.40} & 21.13{\scriptsize ±2.12} & 26.38{\scriptsize ±2.18} & 38.75{\scriptsize ±1.03}\\
OnPro + Ours& \textbf{74.49{\scriptsize ±2.14}} & \textbf{78.64{\scriptsize ±1.42}} & \textbf{34.76{\scriptsize ±1.12}} & \textbf{41.89{\scriptsize ±0.82} }& \textbf{50.01{\scriptsize ±0.85}} & \textbf{21.81{\scriptsize ±1.02}} & \textbf{32.00{\scriptsize ±0.72}} & \textbf{38.18{\scriptsize ±1.02}} & \textbf{47.93{\scriptsize ±1.26}}\\
\bottomrule
\end{tabular}
    }
    \vspace{-8pt}
    \caption{Average Accuracy (\%, higher is better) on four benchmark datasets with difference memory buffer size $M$, with and without our proposed CCL-DC scheme. The result of our method is given by the ensemble of two peer models. All values are averages of 10 runs.}
    \label{tab:aa}
    \vspace{-8pt}
\end{table*}

\begin{table*}
    \centering
    \resizebox{\textwidth}{!}{
    \begin{tabular}{lccccccccc}
\toprule
\multicolumn{1}{c}{Dataset} & \multicolumn{2}{c}{CIFAR10} & \multicolumn{3}{c}{CIFAR100} & \multicolumn{3}{c}{Tiny-ImageNet} & ImageNet-100\\
\cmidrule(lr){1-1}
\cmidrule(lr){2-3}
\cmidrule(lr){4-6}
\cmidrule(lr){7-9}
\cmidrule(lr){10-10}
\multicolumn{1}{c}{Memory Size $M$} & 500 & 1000 & 1000 & 2000 & 5000 & 2000 & 5000 & 10000 & 5000\\
\midrule
ER & 83.13{\scriptsize ±1.60} & 78.15{\scriptsize ±3.60} & 53.77{\scriptsize ±1.51} & 51.53{\scriptsize ±1.66} & 50.79{\scriptsize ±0.71} & 68.15{\scriptsize ±1.47} & 64.99{\scriptsize ±1.22} & 64.44{\scriptsize ±1.45} & 53.95{\scriptsize ±1.51}\\
ER + Ours & \textbf{90.60{\scriptsize ±1.50}} & \textbf{89.99{\scriptsize ±1.50}} & \textbf{72.38{\scriptsize ±0.66}} & \textbf{70.86{\scriptsize ±0.72}} & \textbf{68.84{\scriptsize ±1.05}} & \textbf{85.24{\scriptsize ±0.53}} & \textbf{81.75{\scriptsize ±0.83}} & \textbf{79.54{\scriptsize ±0.74}} & \textbf{68.73{\scriptsize ±1.21}}\\
\midrule
DER++ & 77.14{\scriptsize ±2.96} & 78.00{\scriptsize ±2.16} & 56.13{\scriptsize ±3.75} & 55.33{\scriptsize ±3.26} & 56.32{\scriptsize ±3.44} & 70.01{\scriptsize ±1.83} & 66.87{\scriptsize ±1.30} & 70.28{\scriptsize ±2.42} & 60.65{\scriptsize ±2.97}\\
DER++ + Ours & \textbf{88.85{\scriptsize ±1.88}} & \textbf{89.00{\scriptsize ±1.67}} & \textbf{72.85{\scriptsize ±1.37}} & \textbf{71.54{\scriptsize ±1.99}} & \textbf{69.52{\scriptsize ±2.37}} & \textbf{82.83{\scriptsize ±1.27}} & \textbf{78.80{\scriptsize ±1.62}} & \textbf{77.79{\scriptsize ±0.86}} & \textbf{70.16{\scriptsize ±1.03}}\\
\midrule
ER-ACE & 57.66{\scriptsize ±4.16} & 61.59{\scriptsize ±3.35} & 38.53{\scriptsize ±1.61} & 39.95{\scriptsize ±2.00} & 41.56{\scriptsize ±1.44} & 5.60{\scriptsize ±1.45} & 4.83{\scriptsize ±0.78} & 4.92{\scriptsize ±0.95} & 49.82{\scriptsize ±1.05}\\
ER-ACE + Ours & \textbf{88.37{\scriptsize ±1.39}} & \textbf{88.40{\scriptsize ±1.15}} & \textbf{69.47{\scriptsize ±0.88}} & \textbf{68.39{\scriptsize ±1.32}} & \textbf{66.63{\scriptsize ±0.90}} & \textbf{21.91{\scriptsize ±5.16}} & \textbf{21.88{\scriptsize ±4.39}} & \textbf{18.88{\scriptsize ±3.12}} & \textbf{68.52{\scriptsize ±0.82}}\\
\midrule
OCM & 78.71{\scriptsize ±3.66} & 81.33{\scriptsize ±2.06} & 40.87{\scriptsize ±1.60} & 42.00{\scriptsize ±1.48} & 42.43{\scriptsize ±1.80} & 18.56{\scriptsize ±2.87} & 15.86{\scriptsize ±2.01} & 15.03{\scriptsize ±2.02} & 20.77{\scriptsize ±1.88}\\
OCM + Ours & \textbf{82.39{\scriptsize ±2.23}} & \textbf{84.53{\scriptsize ±1.63}} & \textbf{48.89{\scriptsize ±2.04}} & \textbf{49.83{\scriptsize ±2.01}} & \textbf{49.94{\scriptsize ±2.16}} & \textbf{31.69{\scriptsize ±1.81}} & \textbf{29.54{\scriptsize ±2.35}} & \textbf{28.10{\scriptsize ±2.28}} & \textbf{48.20{\scriptsize ±1.38}}\\
\midrule
GSA & 79.87{\scriptsize ±3.26} & 77.09{\scriptsize ±4.55} & 58.16{\scriptsize ±1.58} & 55.13{\scriptsize ±1.81} & 50.34{\scriptsize ±1.73} & 20.46{\scriptsize ±1.59} & 15.86{\scriptsize ±1.26} & 14.50{\scriptsize ±0.63} & 62.59{\scriptsize ±1.17}\\
GSA + Ours & \textbf{91.69{\scriptsize ±1.11}} & \textbf{90.98{\scriptsize ±1.33}} & \textbf{73.73{\scriptsize ±1.03}} & \textbf{72.68{\scriptsize ±0.98}} & \textbf{70.36{\scriptsize ±1.07}} & \textbf{80.36{\scriptsize ±1.22}} & \textbf{74.77{\scriptsize ±1.66}} & \textbf{70.71{\scriptsize ±1.19}} & \textbf{73.71{\scriptsize ±1.12}}\\
\midrule
OnPro & 84.23{\scriptsize ±2.00} & 85.60{\scriptsize ±1.56} & 41.34{\scriptsize ±1.63} & 42.59{\scriptsize ±1.65} & 42.92{\scriptsize ±1.00} & 20.84{\scriptsize ±1.47} & 16.73{\scriptsize ±1.27} & 15.82{\scriptsize ±1.04} & 39.60{\scriptsize ±0.86}\\
OnPro + Ours & \textbf{90.39{\scriptsize ±1.59}} & \textbf{90.18{\scriptsize ±1.58}} & \textbf{46.30{\scriptsize ±1.10}} & \textbf{47.13{\scriptsize ±1.01}} & \textbf{47.27{\scriptsize ±1.81}} & \textbf{25.87{\scriptsize ±1.91}} & \textbf{21.40{\scriptsize ±1.52}} & \textbf{19.75{\scriptsize ±1.22}} & \textbf{52.55{\scriptsize ±2.18}}\\
\bottomrule
\end{tabular}
    }
    \vspace{-8pt}
    \caption{Learning Accuracy (\%, higher is better) on four benchmark datasets with difference memory buffer size $M$, with and without our proposed CCL-DC scheme. The result of our method is given by the ensemble of two peer models. All values are averages of 10 runs.}
    \label{tab:la}
    \vspace{-1.4em}
\end{table*}

\subsection{Apply CCL-DC to online CL methods}

The overall loss to network $\theta^1$ when adapting CCL-DC can be written as:
\begin{equation}
\begin{aligned}
\mathcal{L}^1 = \mathcal{L}_{Baseline} + \mathcal{L}_{CCL}^1 + \mathcal{L}_{DC}^1,
\end{aligned}
\label{eq:all}
\end{equation}
where $\mathcal{L}_{Baseline}$ is the loss function of the baseline model CCL-DC adapts to. Note that the model $\theta^2$ should be trained similarly. In Algorithm~\ref{code:pseudo_code}, we provide a Pytorch-like pseudo-code demonstrating how to incorporate CCL-DC into a given baseline. For simplicity, we only show the loss function for model $\theta^1$. Also, we omitted the memory buffer in the pseudo-code. However, the training should be consistent with the baseline, using both streaming and memory data.

% In Algorithm~\ref{code:pseudo_code}, we provide a Pytorch-like pseudo-code demonstrating how to incorporate CCL-DC into a given baseline. For simplicity, we only show the loss function for model $\theta^1$, and model $\theta^2$ should be trained similarly. Also, we omitted the memory buffer in the pseudo-code. However, the training of CCL-DC should be consistent with the baseline, using both streaming and memory data.

\section{Experiments}
\label{sec:experiments}

\begin{table*}
    \centering
    \resizebox{\textwidth}{!}{
    \begin{tabular}{lccccccccc}
\toprule
\multicolumn{1}{c}{Dataset} & \multicolumn{2}{c}{CIFAR10} & \multicolumn{3}{c}{CIFAR100} & \multicolumn{3}{c}{Tiny-ImageNet} & ImageNet-100\\
\cmidrule(lr){1-1}
\cmidrule(lr){2-3}
\cmidrule(lr){4-6}
\cmidrule(lr){7-9}
\cmidrule(lr){10-10}
\multicolumn{1}{c}{Memory Size $M$} & 500 & 1000 & 1000 & 2000 & 5000 & 2000 & 5000 & 10000 & 5000\\
\midrule
ER  & 31.63{\scriptsize ±3.81} & 20.63{\scriptsize ±8.32} & 55.71{\scriptsize ±2.24} & 39.11{\scriptsize ±3.87} & 23.05{\scriptsize ±3.69} & 85.00{\scriptsize ±1.30} & 71.62{\scriptsize ±2.18} & 62.43{\scriptsize ±3.83} & 39.26{\scriptsize ±3.21}\\
ER + Ours & \textbf{26.74{\scriptsize ±3.99}} & \textbf{17.58{\scriptsize ±2.71}} & \textbf{54.34{\scriptsize ±2.22}} & \textbf{37.67{\scriptsize ±2.16}} & \textbf{21.98{\scriptsize ±2.59}} & \textbf{81.13{\scriptsize ±1.93}} & \textbf{64.79{\scriptsize ±1.32}} & \textbf{53.18{\scriptsize ±0.99}} & \textbf{37.78{\scriptsize ±2.18}}\\
\midrule
DER++ &23.60{\scriptsize ±3.64} & 17.71{\scriptsize ±2.18} & 55.65{\scriptsize ±4.36} & 41.27{\scriptsize ±4.93} & 31.72{\scriptsize ±3.95} & 87.79{\scriptsize ±2.35} & 73.28{\scriptsize ±3.88} & 72.51{\scriptsize ±5.53} & 42.97{\scriptsize ±5.89}\\
DER++ + Ours &\textbf{22.62{\scriptsize ±3.03}} & \textbf{16.43{\scriptsize ±3.36}} & \textbf{53.45{\scriptsize ±1.40}} & \textbf{39.39{\scriptsize ±2.71}} & \textbf{23.71{\scriptsize ±3.39}} & \textbf{87.16{\scriptsize ±1.60}} & \textbf{73.15{\scriptsize ±2.15}} & \textbf{64.48{\scriptsize ±3.08}} & \textbf{35.32{\scriptsize ±2.80}}\\
\midrule
ER-ACE &\textbf{12.25{\scriptsize ±3.84}} & \textbf{9.92{\scriptsize ±2.83}}  & \textbf{25.88{\scriptsize ±4.10}} & \textbf{17.68{\scriptsize ±1.90}} & \textbf{10.62{\scriptsize ±2.08}} & 57.41{\scriptsize ±2.38} & 44.48{\scriptsize ±1.96} & 37.83{\scriptsize ±3.12} & \textbf{23.92{\scriptsize ±2.05}}\\
ER-ACE + Ours &20.62{\scriptsize ±2.26} & 14.32{\scriptsize ±2.58} & 46.78{\scriptsize ±1.91} & 34.19{\scriptsize ±2.40} & 19.01{\scriptsize ±0.94} & \textbf{56.56{\scriptsize ±4.16}} & \textbf{42.20{\scriptsize ±3.94}} & \textbf{31.13{\scriptsize ±3.52}} & 34.43{\scriptsize ±3.60}\\
\midrule
OCM &13.05{\scriptsize ±4.37} & 11.00{\scriptsize ±3.11} & 31.16{\scriptsize ±2.69} & 17.90{\scriptsize ±3.73} & 6.85{\scriptsize ±2.25} & 56.66{\scriptsize ±2.53} & 40.59{\scriptsize ±1.55} & 30.80{\scriptsize ±2.29} &  \textbf{4.55{\scriptsize ±1.60}}\\
OCM + Ours &\textbf{10.75{\scriptsize ±2.52}} & \textbf{8.45{\scriptsize ±2.63}}  & \textbf{29.65{\scriptsize ±4.00}} & \textbf{17.02{\scriptsize ±3.01}} & \textbf{6.16{\scriptsize ±1.35}} & \textbf{51.58{\scriptsize ±2.81}} & \textbf{35.58{\scriptsize ±2.54}} & \textbf{27.24{\scriptsize ±1.60}} & 15.33{\scriptsize ±2.28}\\
\midrule
GSA &25.02{\scriptsize ±2.83} & \textbf{16.56{\scriptsize ±4.02}} & 53.42{\scriptsize ±3.12} & \textbf{37.29{\scriptsize ±2.60}} & \textbf{20.50{\scriptsize ±4.33}} & \textbf{66.87{\scriptsize ±3.31}} & \textbf{53.42{\scriptsize ±3.84}} & \textbf{43.44{\scriptsize ±3.81}} & \textbf{35.44{\scriptsize ±2.42}}\\
GSA + Ours &\textbf{24.96{\scriptsize ±3.27}} & 16.59{\scriptsize ±2.09} & \textbf{52.29{\scriptsize ±2.06}} & 38.76{\scriptsize ±2.41} & 21.36{\scriptsize ±2.36} & 80.08{\scriptsize ±1.97} & 63.85{\scriptsize ±1.78} & 49.73{\scriptsize ±2.10} & 40.46{\scriptsize ±2.54}\\
\midrule
OnPro &\textbf{16.47{\scriptsize ±4.23}} & 12.93{\scriptsize ±3.02} & 35.03{\scriptsize ±4.45} & 24.26{\scriptsize ±2.31} & 12.04{\scriptsize ±2.11} & 64.69{\scriptsize ±3.36} & 50.47{\scriptsize ±4.20} & 42.81{\scriptsize ±4.63} & \textbf{14.44{\scriptsize ±2.08}}\\
OnPro + Ours &17.54{\scriptsize ±4.15} & \textbf{12.90{\scriptsize ±2.77}} & \textbf{27.64{\scriptsize ±3.29}} & \textbf{17.78{\scriptsize ±1.39}} & \textbf{8.41{\scriptsize ±2.62}} & \textbf{56.03{\scriptsize ±2.96}} & \textbf{38.70{\scriptsize ±1.88}} & \textbf{29.24{\scriptsize ±1.33}} & 15.72{\scriptsize ±3.29}\\
\bottomrule
\end{tabular}

    }
    \vspace{-4pt}
    \caption{Relative Forgetting (\%, lower is better) on four benchmark datasets with difference memory buffer size $M$, with and without our proposed CCL-DC scheme. The result of our method is given by the ensemble of two peer models. All values are averages of 10 runs.}
    \label{tab:rf}
    \vspace{-8pt}
\end{table*}

\subsection{Experimental Setup}
\paragraph{Datasets.}
We use four image classification benchmark datasets to evaluate the effectiveness of our method, 
including CIFAR-10~\cite{krizhevsky2009learning}, CIFAR-100~\cite{krizhevsky2009learning}, TinyImageNet~\cite{le2015tiny}, and ImageNet-100~\cite{hou2019learning, deng2009imagenet, imagenet100pytorch}. More detailed information about the dataset split and task allocation is given in the supplementary material.

\vspace{-10pt}
\paragraph{Baselines.}
To show the effectiveness of our strategy, we applied CCL-DC to six typical and state-of-the-art online CL methods, including ER~\cite{rolnick2019experience}, DER++~\cite{buzzega2020dark}, ER-ACE~\cite{caccia2021new}, OCM~\cite{guo2022online}, GSA~\cite{guo2023dealing}, and OnPro~\cite{wei2023online}. 

\vspace{-10pt}
\paragraph{Implementation details.} 
We use full ResNet-18~\cite{he2016deep} (not pre-trained) as the backbone for every method. For each baseline method, we perform a hyperparameter search on CIFAR-100, $M=$2k, and apply the hyperparameter to all of the settings. For fair comparison, we use the same optimizer and hyperparameter setting when adapting CCL-DC to the baselines. For hyperparameters unique to CCL-DC, we conduct another hyperparameter search as stated in the supplementary material. We set the streaming batch size to 10 and the memory batch size to 64. We do not use the multiple update trick as described in~\cite{NIPS2019_9357}. 
More detailed information about data augmentation, hyperparameter search, and hardware environments is given in the supplementary material.

\subsection{Results and Analysis}
\paragraph{Final average accuracy.} 
Table~\ref{tab:aa} presents average accuracy (AA) at the end of the training on four datasets. As indicated in Sec.~\ref{sec:method}, to fully take advantage of collaborative learning, we show the results with the ensemble of two models, with the independent model performance available in the supplementary material. Generally, the ensemble method provides about 1\% additional accuracy compared to independent inference. For all datasets, memory size $M$, and baseline methods, applying CCL-DC constantly improves the performance by a large margin. Notably, even for state-of-the-art methods such as GSA and OnPro, we can still gain significant performance when incorporating CCL-DC. 

More interestingly, for almost all settings with different memory buffer sizes $M$, the performance gain tends to be a constant on a relative basis. For example, CCL-DC can boost the performance of ER on Tiny-ImageNet from 10.82 to 16.56 when $M$=2k, which is a 53.0\% performance gain on a relative basis. The performance gain is 53.4\% and 52.7\% when $M$=5k and $M$=10k, respectively. This indicates that we can achieve a decent performance gain regardless of the memory buffer size, and it shows the scalability of our method to different resource conditions.

\begin{table}
    \centering
    \resizebox{0.75\linewidth}{!}{
    \begin{tabular}{lcc}
\toprule
\multicolumn{1}{c}{Method} & Acc. ↑ & LA ↑ \\
\midrule
ER & 31.89{\scriptsize ±1.45} & 51.53{\scriptsize ±1.66}\\
ER + Multivew & 38.18{\scriptsize ±1.46} & 64.02{\scriptsize ±1.12}\\
ER + Ours (CCL only)& 41.05{\scriptsize ±1.21} & 68.76{\scriptsize ±0.79}\\
ER + Ours & 44.45{\scriptsize ±1.04} & 70.86{\scriptsize ±0.72}\\
\midrule
ER-ACE & 34.21{\scriptsize ±1.53} & 39.95{\scriptsize ±2.00}\\
ER-ACE + Multivew & 38.61{\scriptsize ±1.48} & 47.45{\scriptsize ±1.88}\\
ER-ACE + Ours (CCL only) & 40.90{\scriptsize ±1.08} & 50.91{\scriptsize ±1.63}\\
ER-ACE + Ours & 45.14{\scriptsize ±1.00} & 68.39{\scriptsize ±1.32}\\
\bottomrule
\end{tabular}
    }
    \caption{Ablation studies on CIFAR-100 (M=2k). We report the ensemble performance for methods incorporating CCL.}
    \vspace{-12pt}
    \label{tab:ablation}
\end{table}

\vspace{-10pt}
\paragraph{Plasticity and stability metric.}
As mentioned in Sec.~\ref{sec:tradeoff}, we evaluate the plasticity and stability of different continual learners with LA and RF, respectively. Table~\ref{tab:la} shows the plasticity metric on four datasets. For all settings, CCL-DC constantly improves the model plasticity by a large margin. For model stability, as indicated by RF in Table~\ref{tab:rf}, models trained with CCL-DC are comparable with the baselines under most cases. ER-ACE is an exception as its plasticity is unexpectedly low, especially on TinyImagenet. Also, the stability of ER-ACE is compromised when incorporating CCL-DC. We will explain the reason for this unexpected phenomenon in the supplementary material.
% \paragraph{Confusion Matrices at the end of training.}

\subsection{Ablation Studies}
\paragraph{Effect of multiview learning.}
As mentioned in Sec.~\ref{sec:method}, CCL-DC benefits from multiview learning with data augmentation in DC. For fair comparison, we explore how multiview learning will impact the performance of the baselines. We apply the classification loss part of CCL-DC to the baselines. Table~\ref{tab:ablation} demonstrates that multiview learning can improve both AA and LA of baselines. However, those performance gains are still inferior to CCL-DC. 

\vspace{-10pt}
\paragraph{Effect of CCL.} 
We evaluate how CCL alone can improve the baselines. In the experiments, we remove multiview learning and DC, and we train the continual learner pair with the loss illustrated in Eq.~\ref{eq:CCL}. Table~\ref{tab:ablation} shows the performance gain for ER and ER-ACE. We can see that CCL alone can provide significant gains in both final accuracy and plasticity. Also, when combining CCL with DC, the performance can be further improved. 

\vspace{-10pt}
\paragraph{Distillation scheme of DC.}
We also evaluate the effectiveness of DC's strategy of distilling from harder samples to easier samples in collaborative learning manner. As shown in Table~\ref{tab:ablation_dc}, we compared it with other distillation strategies. The result shows that the distillation scheme of DC constantly outperforms other schemes.

\vspace{-10pt}
\paragraph{Self-Distillation Chain in an independent model.}
While DC was originally designed to incorporate CCL, it is possible to conduct DC's strategy within an independent model. We name such a strategy as the Self-Distillation Chain (SDC). Similar to different schemes of DC, SDC can be implemented in two ways: distillation from easier samples to harder samples, and distillation from harder samples to easier samples. As shown in Table~\ref{tab:rc}, both strategy gives extra final performance and plasticity, while the latter strategy benefits the performance more. Moreover, incorporating DC with CCL (i.e., ours) further improves the accuracy. 

% \paragraph{Robustness w.r.t. augmentation.}

\begin{table}
    \centering
    \resizebox{0.8\linewidth}{!}{
    \begin{tabular}{clcc}
\toprule
Method & \multicolumn{1}{c}{Distillation scheme} & Acc. ↑ & LA ↑\\
\midrule
ER & Easy to hard & 40.95{\scriptsize ±0.97} & 60.03{\scriptsize ±0.98}\\
ER & Same difficulty & 43.64{\scriptsize ±1.09} & 69.49{\scriptsize ±0.78}\\
ER & Hard to easy (Ours) & 44.45{\scriptsize ±1.04} & 70.86{\scriptsize ±0.72}\\
\midrule
ER-ACE & Easy to hard & 38.46{\scriptsize ±1.51} & 39.00{\scriptsize ±1.03} \\
ER-ACE & Same difficulty & 43.81{\scriptsize ±1.28} & 55.37{\scriptsize ±1.54} \\
ER-ACE & Hard to easy (Ours) & 45.14{\scriptsize ±1.00} & 68.39{\scriptsize ±1.32}  \\
\bottomrule
\end{tabular}
    } 
    \caption{Comparison of different distillation schemes in DC on CIFAR-100 (M=2k).}
    \vspace{-8pt}
    \label{tab:ablation_dc}
\end{table}

\begin{table}
    \centering
    \resizebox{0.87\linewidth}{!}{
    \begin{tabular}{clcc}
    \toprule
    Method & \multicolumn{1}{c}{Distillation scheme} & Acc. ↑ & LA ↑\\
    \midrule
    ER + SDC & Easy to hard & 35.00{\scriptsize ±1.31} & 56.67{\scriptsize ±0.84}\\
    ER + SDC & Hard to easy & 41.31{\scriptsize ±1.25} & 68.62{\scriptsize ±0.60}\\
    \midrule
    ER + CCL-DC & Easy to hard & 40.95{\scriptsize ±0.97} & 60.03{\scriptsize ±0.98}\\
    ER + CCL-DC & Hard to easy (Ours) & 44.45{\scriptsize ±1.04} & 70.86{\scriptsize ±0.72}\\
    \bottomrule
    \end{tabular}
    }
    \vspace{-.5em}
    \caption{Final average accuracy when conducting multiview distillation within the same model on CIFAR-100 (M=2k). 
    All values are averaged over 10 runs.}
    \vspace{-1em}
    \label{tab:rc}
\end{table}
\section{Discussions}
\label{sec:discussions}
\begin{figure}[t]
  \centering
   \includegraphics[width=0.75\linewidth]{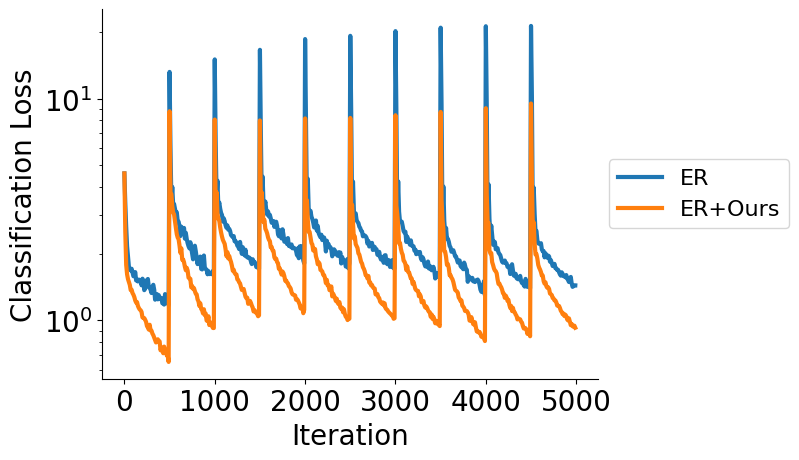}
   \vspace{-8pt}
   \caption{Classification loss curve of ER on CIFAR-100 (M=2k). The curve is calculated on all training samples of the \textit{current} task. Since there are 10 tasks in total, the curve has 10 peaks.}
   \vspace{-1.4em}
   \label{fig:CE}
\end{figure}

In this section, we analyze some properties of CCL-DC. 
\vspace{-10pt}
\paragraph{Improving plasticity.}
One of the important advantages of CCL-DC is that it can improve the plasticity of continual learners. This can be evident by plasticity metrics like LA. Moreover, we have observed that the plasticity of CCL-DC facilitates the model to converge faster and descend to a deeper loss. Figure \ref{fig:CE} illustrates the classification loss (cross-entropy) curve of the model. To obtain the loss curve, we take a snapshot of the model every 10 iterations and compute the cross-entropy over all the training samples on the \textit{current} task. We plot the curve on the logarithm scale so that it is easy to observe that CCL-DC helps the model descend deeper at the end of each task. 

\begin{figure}[t]
  \centering
   \includegraphics[width=0.65\linewidth]{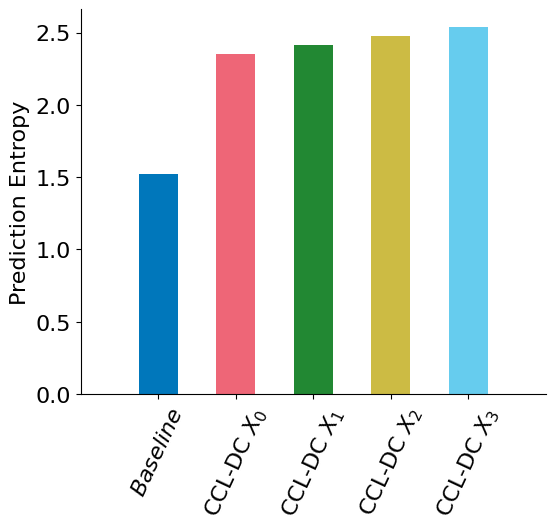}
   \vspace{-8pt}
   \caption{The entropy of prediction produced by ER with and without CCL-DC on CIFAR-100 (M=2k). $X_i$ represents the sample after $i$-th augmentation in Eq.~\ref{eq:DC}. The value is calculated at the end of the training and is averaged over all training samples.}
   \label{fig:entropy}
   \vspace{-1.4em}
\end{figure}

\vspace{-10pt}
\paragraph{Alleviating overconfidence.}
To show the effect of DC in alleviating overconfidence, we compared the confidence of models trained with and without CCL-DC. We measure the confidence with the entropy of predictions. Fig.~\ref{fig:entropy} shows the entropy of prediction produced by the ER (baseline) with and without CCL-DC on CIFAR-100 (M=2k). The entropy is calculated at the end of training and averaged across all samples in the training set. Moreover, we also calculate the entropy of models trained with CCL-DC, by forwarding images of different difficulties (from $X_0$ to $X_3$ in Eq.~\ref{eq:DC}) in the augmentation chain. The experimental results show the effect of DC in suppressing overall confidence.

\section{Conclusion}
\label{sec:conclusion}
In this paper, we highlighted the significance of plasticity in online CL, which has been overlooked in prior research when compared to stability. We also established the quantitative link between plasticity, stability, and final accuracy. The quantitative relationship shed light on the future direction of online CL research. Based on this, we introduced collaborative learning into online CL and proposed CCL-DC, a strategy that can be seamlessly integrated into existing online CL methods. Extensive experiments showed the effectiveness of CCL-DC in boosting plasticity and subsequently improving the final performance. 
{
    \small
    \bibliographystyle{ieeenat_fullname}
    \bibliography{main}
}

\clearpage
\maketitlesupplementary

\section{Extra Experiments}
\label{sec:extra}

\begin{figure}[t]
  \centering
   \includegraphics[width=0.7\linewidth]{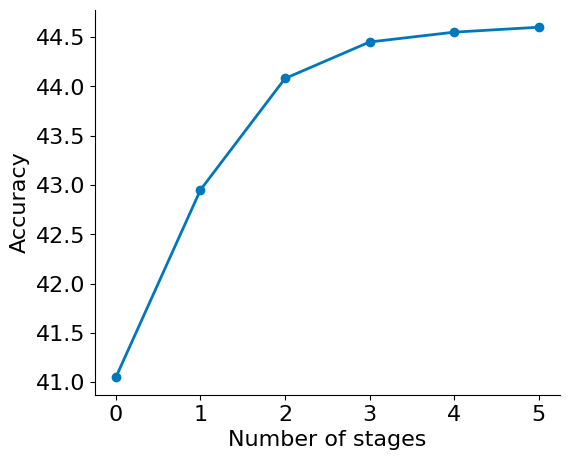}
   \caption{The performance of ER incorporating CCL-DC with varying numbers of augmentation stages on CIFAR-100 ($M=$2k). All numbers are averaged over 10 runs.}
   \label{fig:stages}
\end{figure}

\paragraph{Impact of the number of augmentation stages in DC.} As mentioned in Sec.~\ref{sec:method}, DC comprises three augmentation stages, including one geometric distortion stage and two RandAugment stages. In this ablation study, we aim to investigate how more or less augmentation stages will impact the final performance. In this experiment, we generalize the number of augmentation stages in CCL-DC from 0 (which is equivalent to CCL without DC) to $N$. For $N \ge 1$, we apply one geometric distortion stage and $N-1$ RandAugment stages. As shown in Fig.~\ref{fig:stages}, CCL-DC performs better when the number of augmentation stages increases. However, the training time and memory footprint also increase with more augmentation stages. Thus, for a trade-off, we set the number of stages to 3 in the main paper.

\begin{figure*}
\begin{center}
    \subfloat[ER]{
       \includegraphics[width=0.24\linewidth]{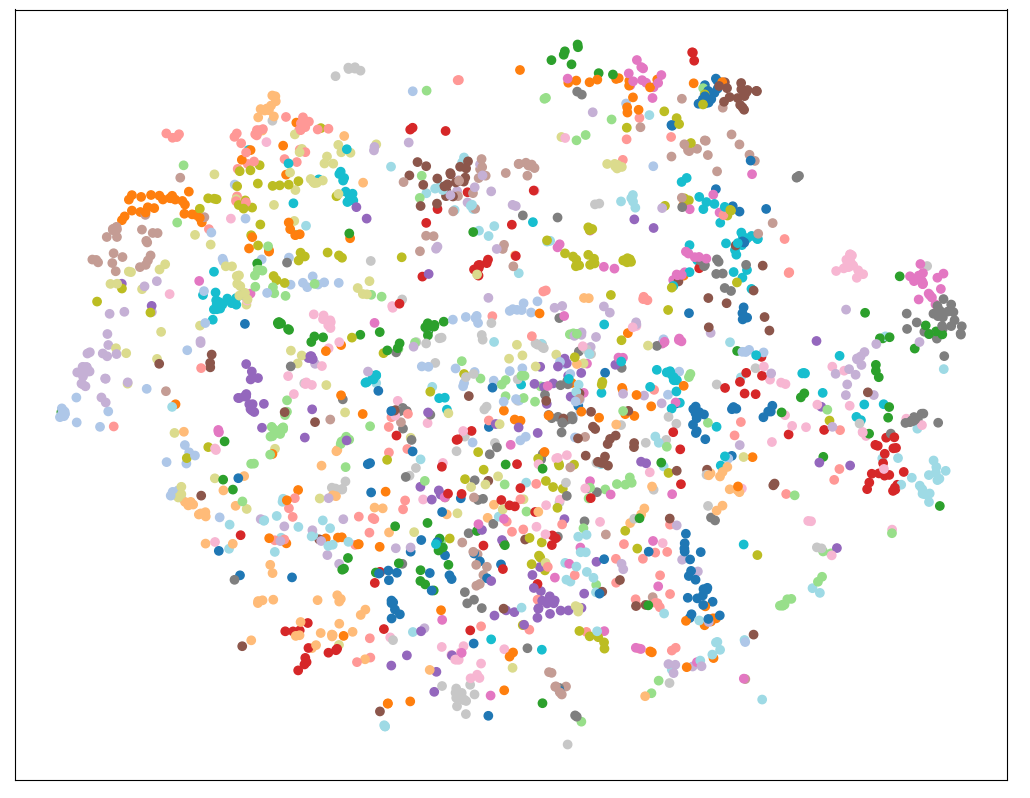}}
    \label{1a}\hfill
    \subfloat[ER + Ours]{
        \includegraphics[width=0.24\linewidth]{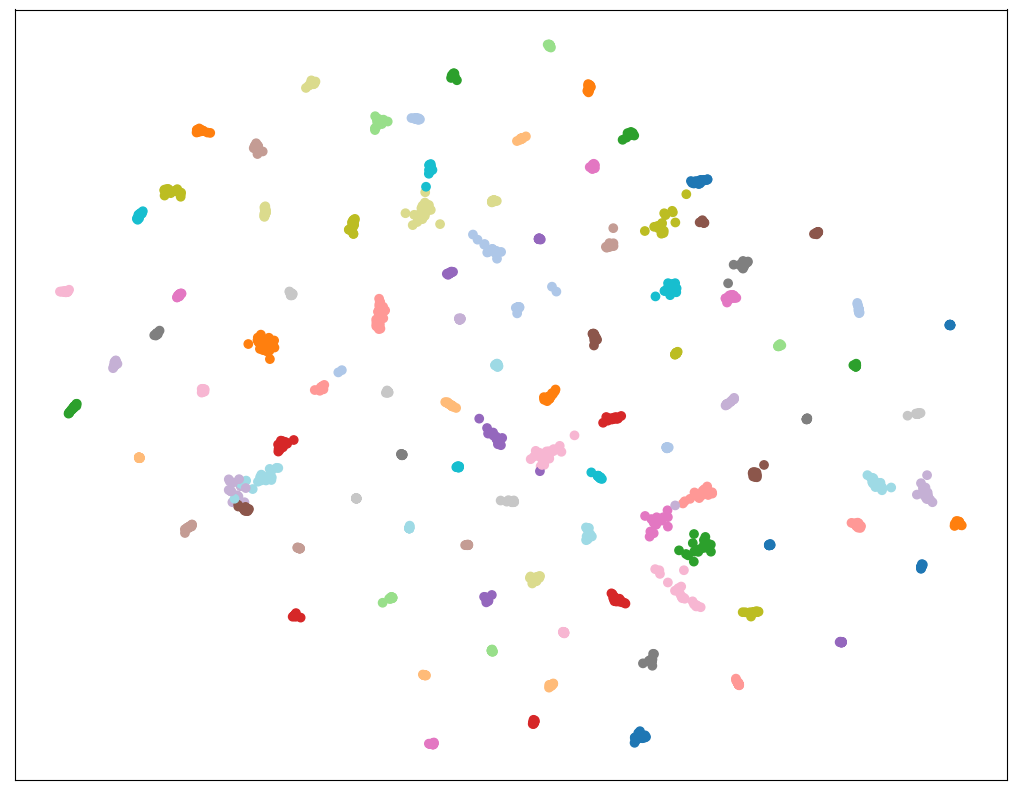}}
    \label{1b}\hfill
        \subfloat[DER++]{
       \includegraphics[width=0.24\linewidth]{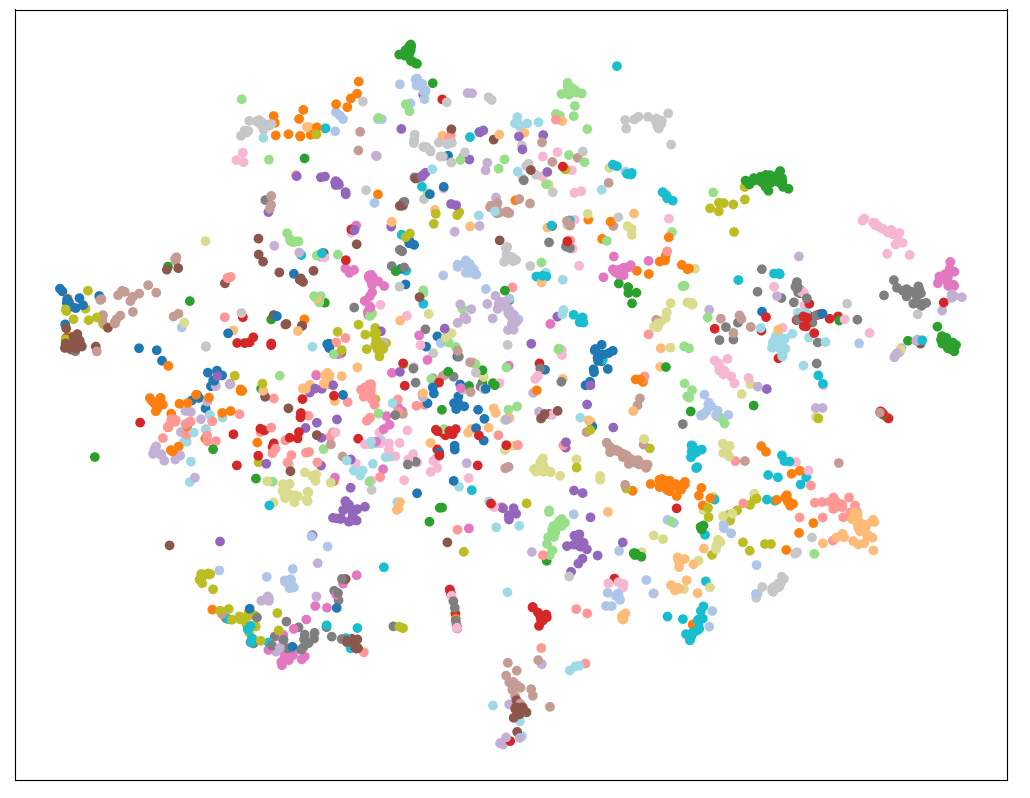}}
    \label{2a}\hfill
    \subfloat[DER++ + Ours]{
        \includegraphics[width=0.24\linewidth]{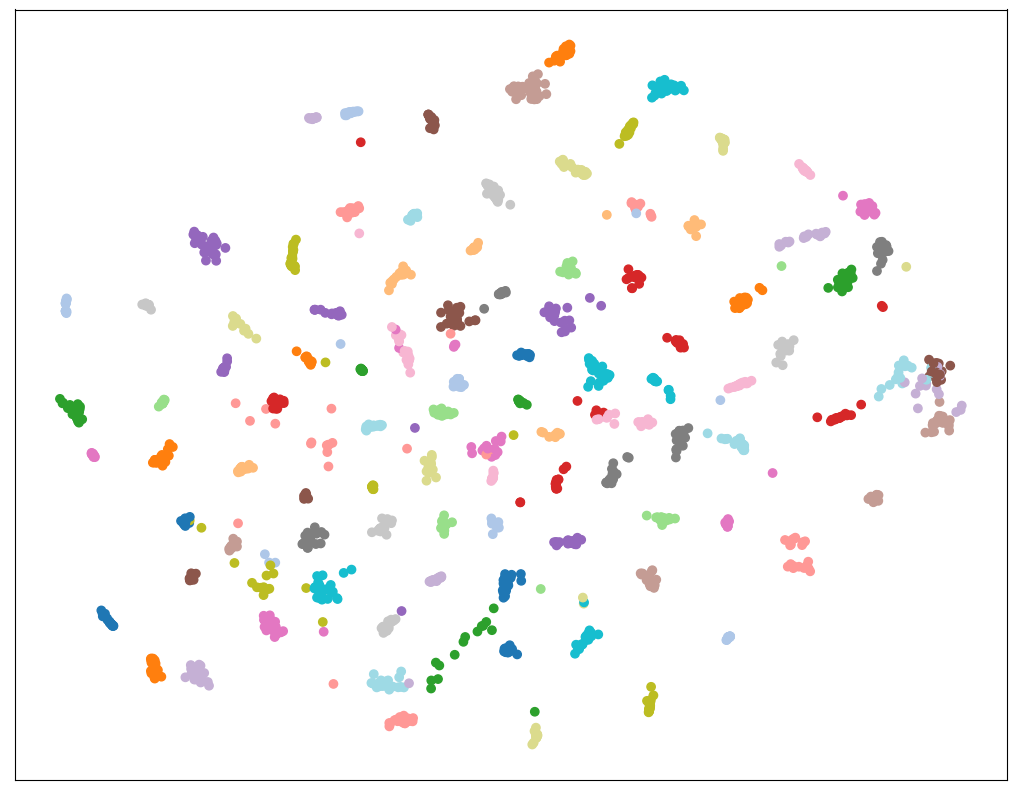}}
    \label{2b}\\
    \subfloat[ER-ACE]{
       \includegraphics[width=0.24\linewidth]{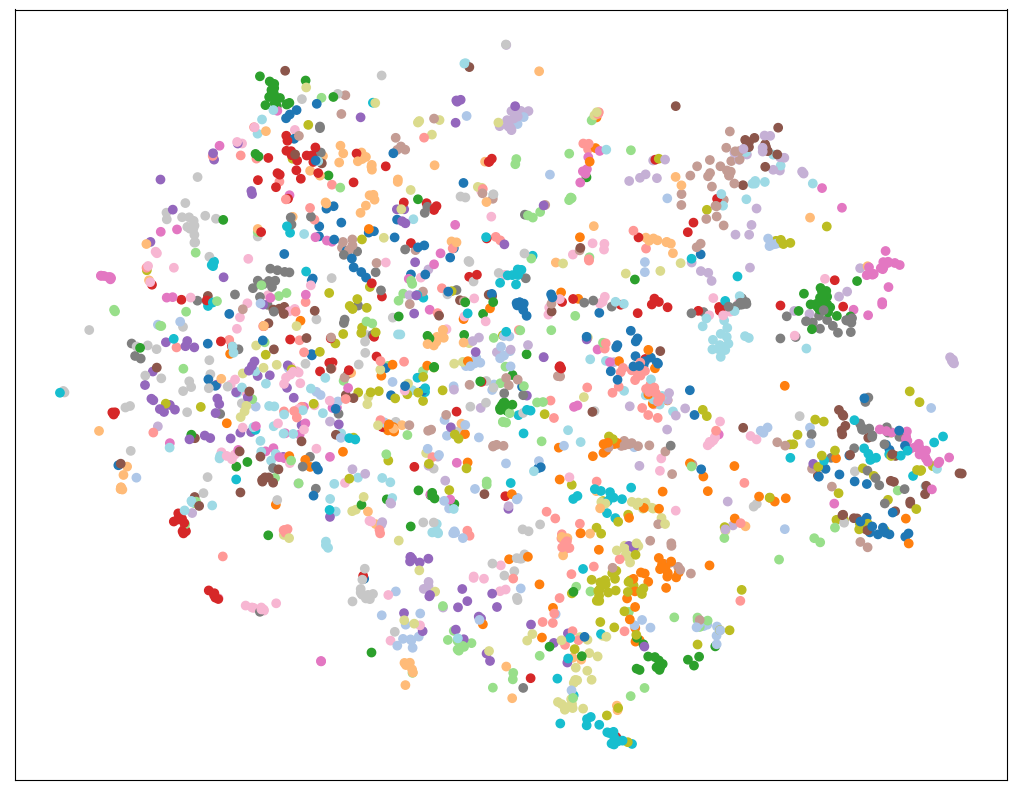}}
    \label{3a}\hfill
    \subfloat[ER-ACE + Ours]{
        \includegraphics[width=0.24\linewidth]{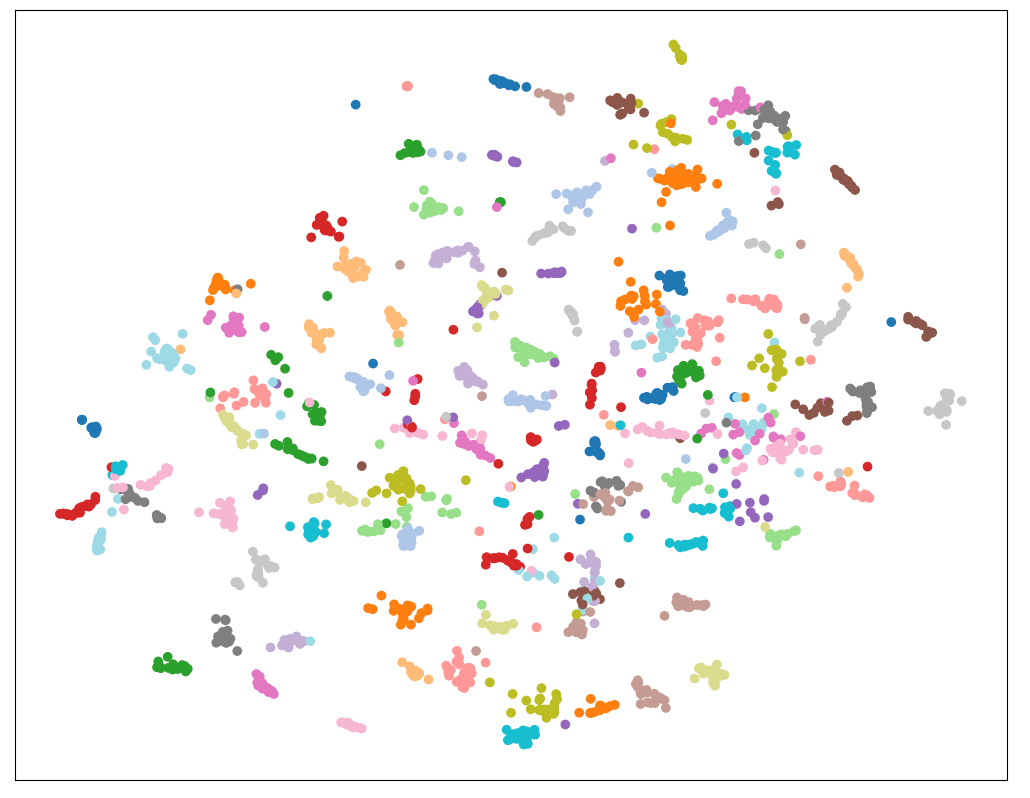}}
    \label{3b}\hfill
    \subfloat[OCM]{
       \includegraphics[width=0.24\linewidth]{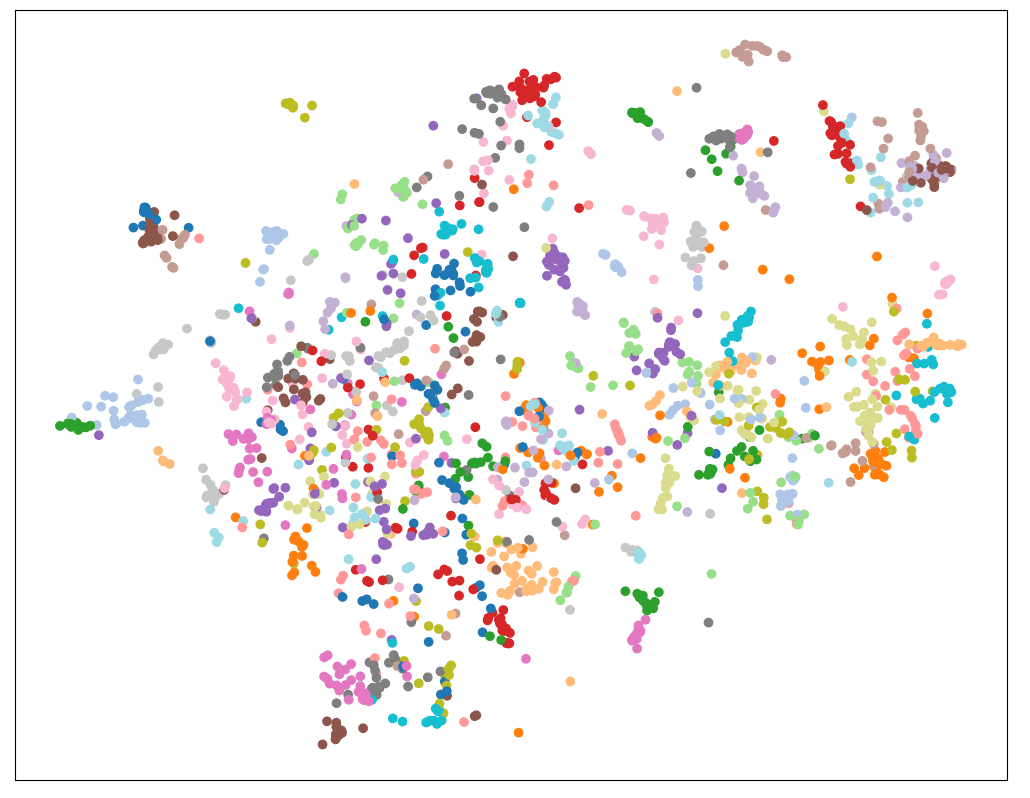}}
    \label{4a}\hfill
    \subfloat[OCM + Ours]{
        \includegraphics[width=0.24\linewidth]{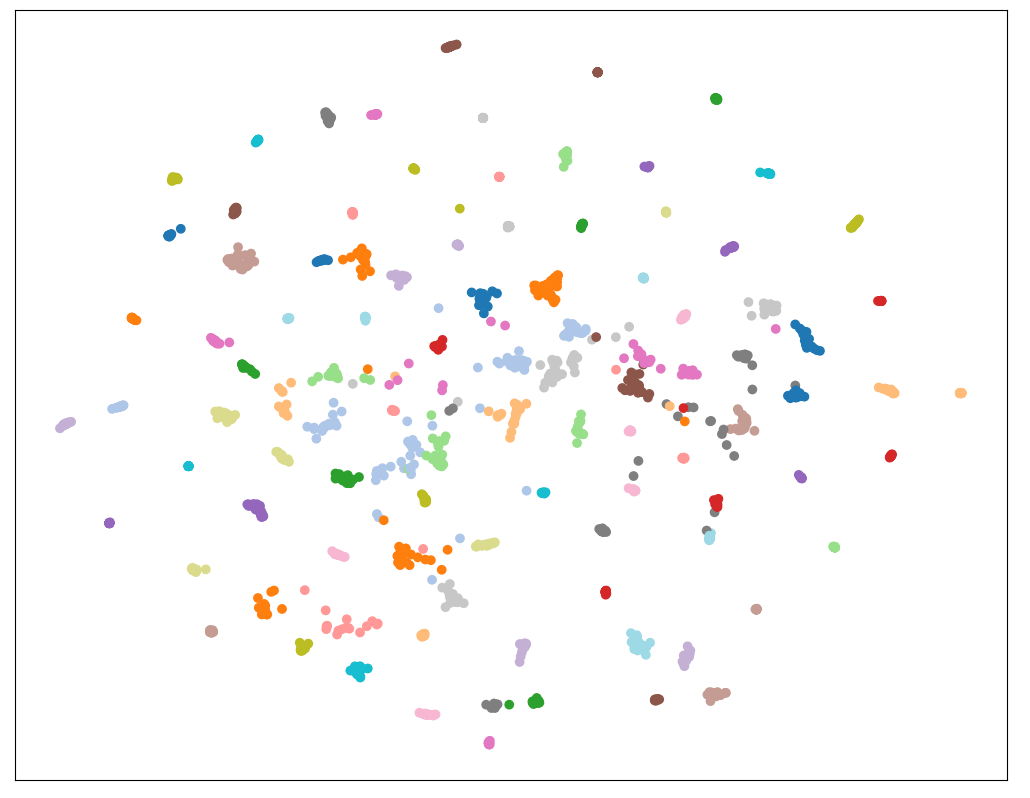}}
    \label{4b} \\
    \subfloat[GSA]{
       \includegraphics[width=0.24\linewidth]{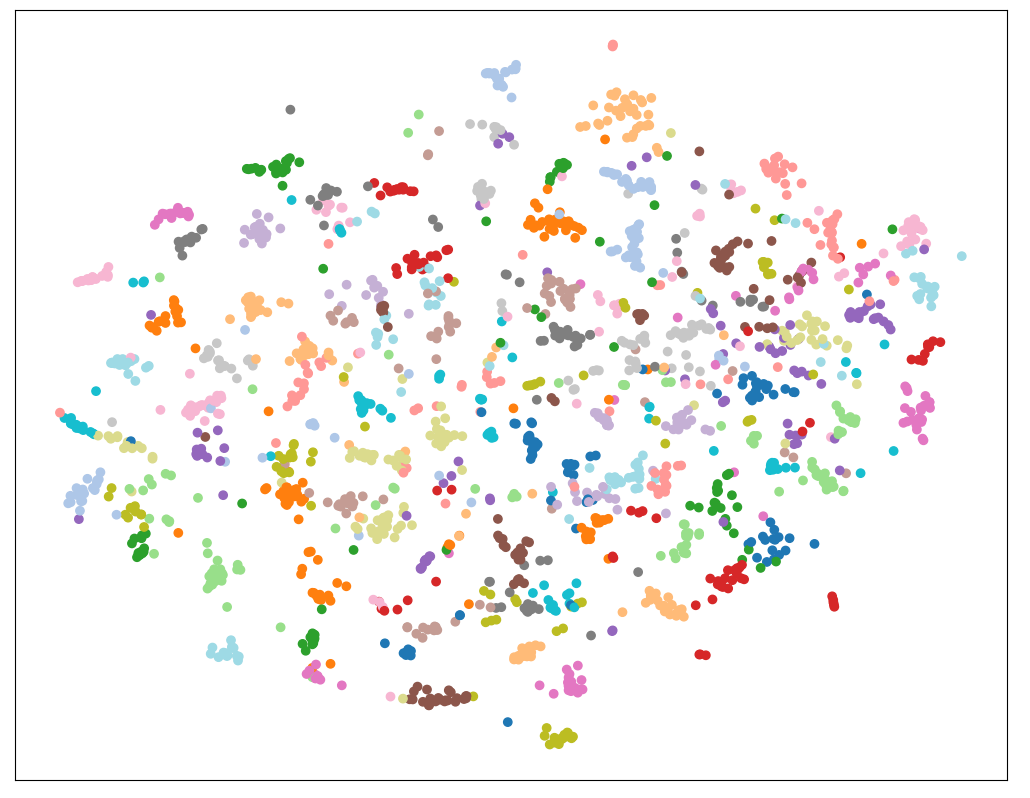}}
    \label{3a}\hfill
    \subfloat[GSA + Ours]{
        \includegraphics[width=0.24\linewidth]{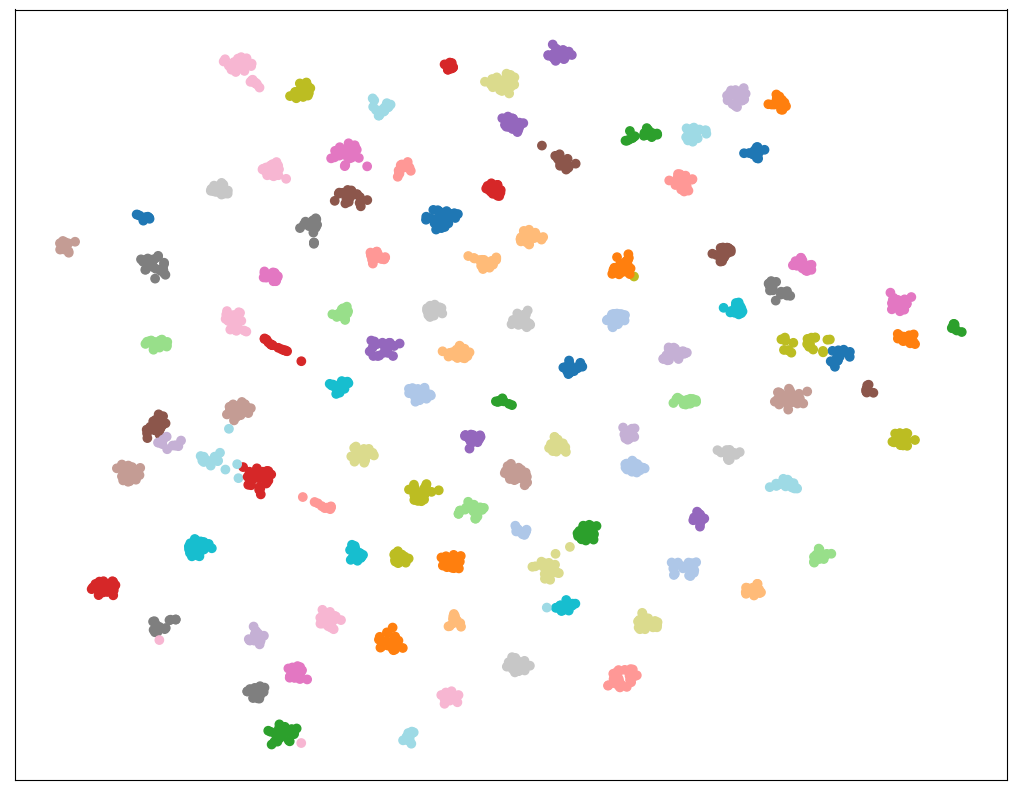}}
    \label{3b}\hfill
    \subfloat[OnPro]{
       \includegraphics[width=0.24\linewidth]{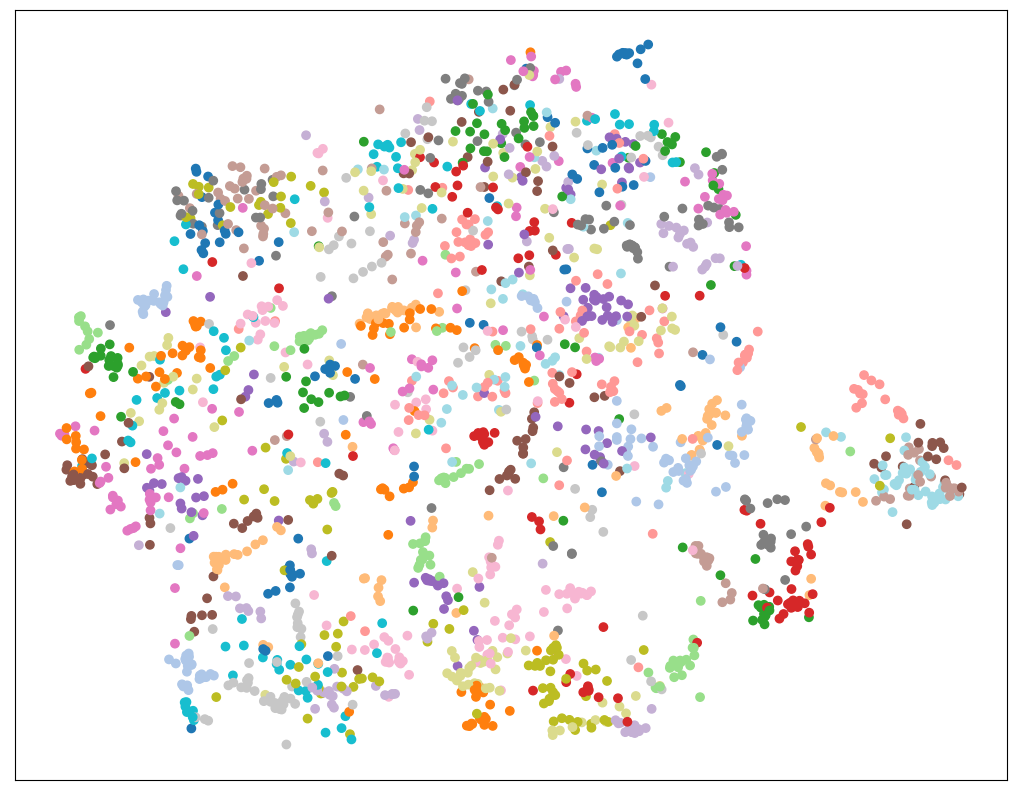}}
    \label{4a}\hfill
    \subfloat[OnPro + Ours]{
        \includegraphics[width=0.24\linewidth]{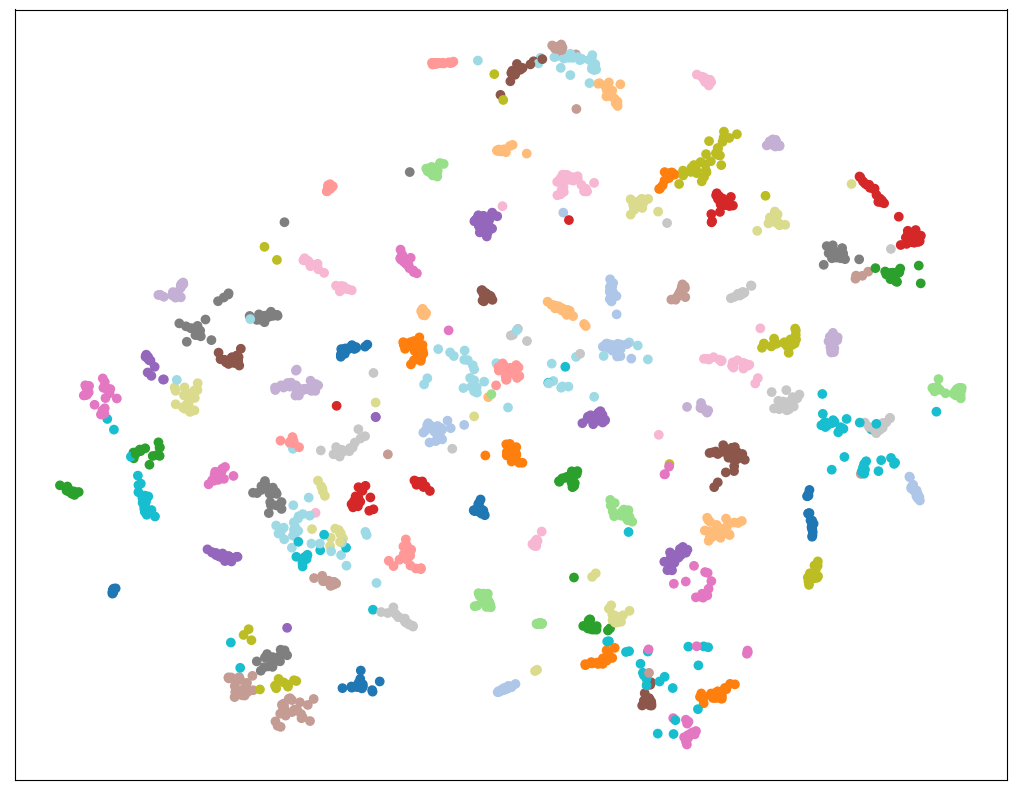}}
    \label{4b} \\
   \vspace{-8pt}
    \caption{T-SNE visualization of memory data at the end of training on CIFAR-100 ($M=$2k).}
    \label{fig:t-SNE_extra} 
    \vspace{-18pt}
\end{center}
\end{figure*}

\paragraph{T-SNE visualization.} 
Another advantage of CCL-DC is its ability to enhance the feature discrimination of continual learners. Fig.~\ref{fig:t-SNE_extra} illustrates the t-SNE visualization of the memory data's embedding space at the end of the training. We can see that the feature representation of the method with CCL-DC is more discriminative compared with the baseline.

\begin{figure*}[t]
    \subfloat{
       \includegraphics[width=0.32\linewidth]{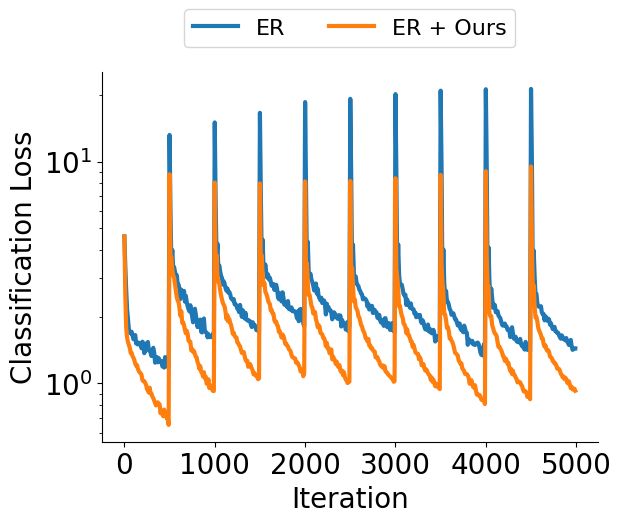}}
    \hfill
    \subfloat{
       \includegraphics[width=0.32\linewidth]{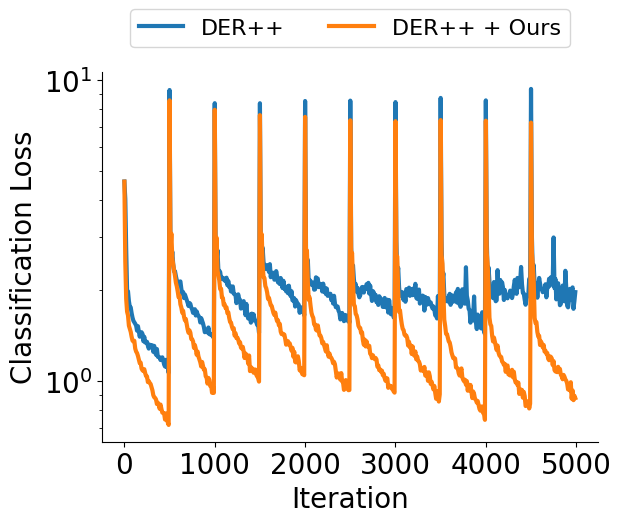}}
    \hfill
    \subfloat{
       \includegraphics[width=0.32\linewidth]{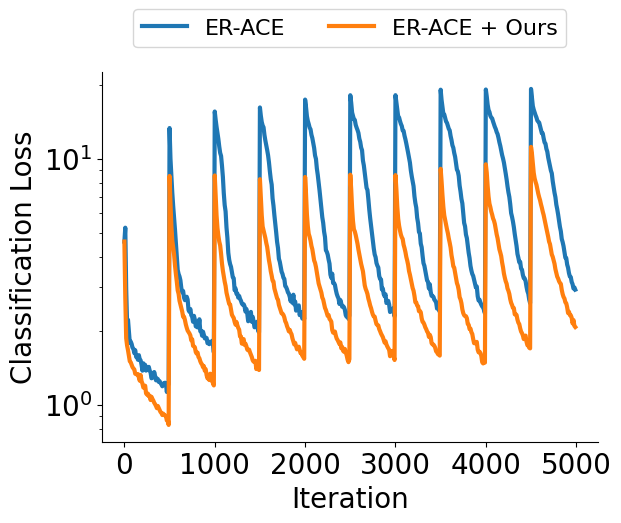}}
    \\
    \subfloat{
       \includegraphics[width=0.32\linewidth]{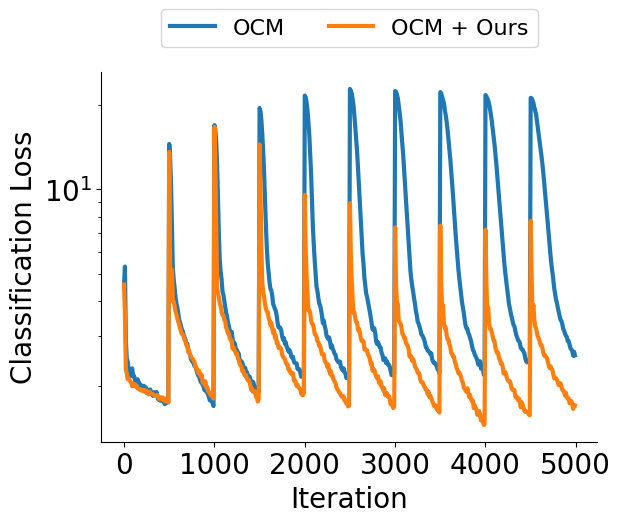}}
    \hfill
    \subfloat{
       \includegraphics[width=0.32\linewidth]{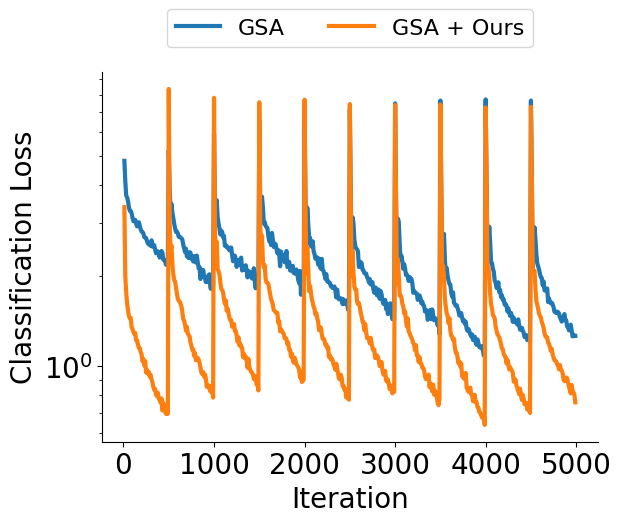}}
    \hfill
    \subfloat{
       \includegraphics[width=0.32\linewidth]{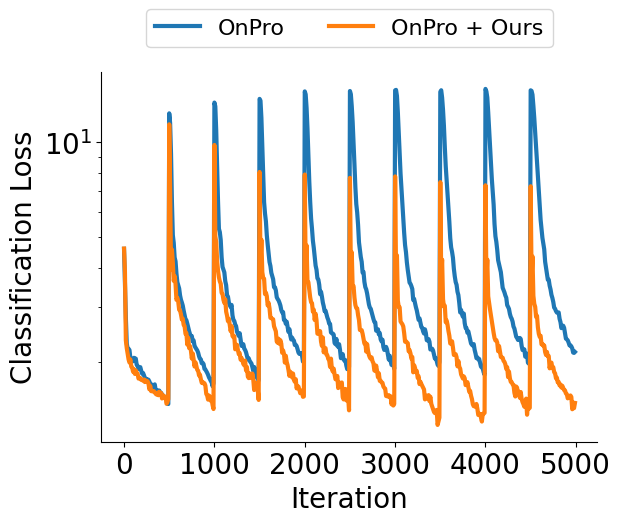}}
    \\
   \vspace{-10pt}
   \caption{Classification loss curve on CIFAR-100 ($M=$2k). The curve is calculated on all training samples of the \textit{current} task. Since there are 10 tasks in total, the curve has 10 peaks.}
   \vspace{-8pt}
   \label{fig:CE_extra}
\end{figure*}

\paragraph{Classification loss curve on other baselines.} In Sec.~\ref{sec:discussions}, we present the classification loss curve of the model during training and illustrate how CCL-DC can assist the model in descending deeper into the loss landscape. We present the classification curve for the remaining baselines in Fig.~\ref{fig:CE_extra}. With improved plasticity, for every baseline method, CCL-DC can improve the training by descending deeper at the end of each task.

\paragraph{Independent network performance.}
Although the ensemble method gives extra performance at inference time, by averaging the logit output of two networks in CCL-DC, it also doubles the computation. In some cases, computational efficiency becomes more crucial during inference. Continual learners trained with CCL-DC are also able to do inference independently, albeit with a slight performance drop compared with ensemble inference. Table~\ref{tab:indaa} shows the accuracy achieved through independent inference. It is evident that the performance loss in independent inference, when compared to ensemble inference, is minimal (approximately 1\%).

\begin{table*}
    \centering
    \resizebox{\textwidth}{!}{
    \begin{tabular}{lccccccccc}
\toprule
\multicolumn{1}{c}{Dataset} & \multicolumn{2}{c}{CIFAR10} & \multicolumn{3}{c}{CIFAR100} & \multicolumn{3}{c}{Tiny-ImageNet} & ImageNet-100\\
\cmidrule(lr){1-1}
\cmidrule(lr){2-3}
\cmidrule(lr){4-6}
\cmidrule(lr){7-9}
\cmidrule(lr){10-10}
\multicolumn{1}{c}{Memory Size $M$} & 500 & 1000 & 1000 & 2000 & 5000 & 2000 & 5000 & 10000 & 5000\\
\midrule
ER + Ours~(Ind.) & 65.66{\scriptsize ±2.35} & 73.37{\scriptsize ±1.70} & 32.97{\scriptsize ±1.06} & 43.58{\scriptsize ±1.05} & 52.96{\scriptsize ±1.16} & 16.32{\scriptsize ±1.58} & 28.68{\scriptsize ±1.20} & 37.14{\scriptsize ±0.93} & 41.82{\scriptsize ±1.54} \\
ER + Ours~(Ens.) & \textbf{66.43{\scriptsize ±2.48}} & \textbf{74.10{\scriptsize ±1.71}} & \textbf{33.43{\scriptsize ±1.06}} & \textbf{44.45{\scriptsize ±1.04}} & \textbf{53.81{\scriptsize ±1.16}} & \textbf{16.56{\scriptsize ±1.63}} & \textbf{29.39{\scriptsize ±1.23}} & \textbf{37.73{\scriptsize ±0.85}} & \textbf{43.11{\scriptsize ±1.49}}\\
\midrule
DER++ + Ours~(Ind.) & 68.15{\scriptsize ±1.40} & 73.56{\scriptsize ±1.12} & 33.81{\scriptsize ±0.90} & 42.79{\scriptsize ±1.38} & 52.04{\scriptsize ±0.81} & \textbf{11.11{\scriptsize ±1.53}} & 21.47{\scriptsize ±1.93} & 27.37{\scriptsize ±2.64} & 44.22{\scriptsize ±1.25} \\
DER++ + Ours~(Ens.) & \textbf{68.79{\scriptsize ±1.42}} & \textbf{74.25{\scriptsize ±1.10}} & \textbf{34.36{\scriptsize ±0.89}} & \textbf{43.52{\scriptsize ±1.35}} & \textbf{52.95{\scriptsize ±0.86}} & 10.99{\scriptsize ±1.39} & \textbf{21.68{\scriptsize ±1.94}} & \textbf{28.01{\scriptsize ±2.46}} & \textbf{45.70{\scriptsize ±1.32}}\\
\midrule
ER-ACE + Ours~(Ind.) & 69.35{\scriptsize ±1.24} & 74.86{\scriptsize ±1.06} & 36.34{\scriptsize ±1.08} & 44.15{\scriptsize ±1.05} & 52.94{\scriptsize ±0.44} & 17.99{\scriptsize ±1.56} & 25.69{\scriptsize ±2.00} & 31.69{\scriptsize ±1.69} & 43.92{\scriptsize ±1.71} \\
ER-ACE + Ours~(Ens.)& \textbf{70.08{\scriptsize ±1.38}} & \textbf{75.56{\scriptsize ±1.14}} & \textbf{37.20{\scriptsize ±1.15}} & \textbf{45.14{\scriptsize ±1.00}} & \textbf{53.92{\scriptsize ±0.48}} & \textbf{18.32{\scriptsize ±1.49}} & \textbf{26.22{\scriptsize ±2.01}} & \textbf{32.23{\scriptsize ±1.70}} & \textbf{45.15{\scriptsize ±1.94}}\\
\midrule
OCM + Ours~(Ind.) & 73.00{\scriptsize ±0.88} & 76.66{\scriptsize ±1.38} & 34.02{\scriptsize ±1.22} & 42.39{\scriptsize ±1.36} & 50.19{\scriptsize ±1.36} & 22.53{\scriptsize ±1.28} & 32.16{\scriptsize ±0.96} & 38.02{\scriptsize ±0.94} & 41.71{\scriptsize ±1.07}  \\
OCM + Ours~(Ens.) & \textbf{74.14{\scriptsize ±0.85}} & \textbf{77.66{\scriptsize ±1.46}} & \textbf{35.00{\scriptsize ±1.15}} & \textbf{43.34{\scriptsize ±1.51}} & \textbf{51.43{\scriptsize ±1.37}} & \textbf{23.36{\scriptsize ±1.18}} & \textbf{33.17{\scriptsize ±0.97}} & \textbf{39.25{\scriptsize ±0.88}} & \textbf{43.19{\scriptsize ±0.98}}\\
\midrule
GSA + Ours~(Ind.) & 68.10{\scriptsize ±1.58} & 74.78{\scriptsize ±1.27} & 35.14{\scriptsize ±1.40} & 43.84{\scriptsize ±1.34} & 54.29{\scriptsize ±1.10} & 16.53{\scriptsize ±1.62} & 27.57{\scriptsize ±1.61} & 36.12{\scriptsize ±1.59} & 43.27{\scriptsize ±1.05}  \\
GSA + Ours~(Ens.) & \textbf{68.91{\scriptsize ±1.68}} & \textbf{75.78{\scriptsize ±1.16}} & \textbf{35.56{\scriptsize ±1.39}} & \textbf{44.74{\scriptsize ±1.32}} & \textbf{55.39{\scriptsize ±1.09}} & \textbf{16.70{\scriptsize ±1.66}} & \textbf{28.11{\scriptsize ±1.70}} & \textbf{37.13{\scriptsize ±1.75}} & \textbf{44.28{\scriptsize ±1.16}}\\
\midrule
OnPro + Ours~(Ind.) & 73.65{\scriptsize ±2.16} & 77.84{\scriptsize ±1.33} & 34.20{\scriptsize ±1.12} & 41.18{\scriptsize ±0.83} & 49.18{\scriptsize ±0.81} & 21.22{\scriptsize ±1.05} & 31.13{\scriptsize ±0.71} & 37.30{\scriptsize ±0.93} & 46.84{\scriptsize ±1.33}  \\
OnPro + Ours~(Ens.) & \textbf{74.49{\scriptsize ±2.14}} & \textbf{78.64{\scriptsize ±1.42}} & \textbf{34.76{\scriptsize ±1.12}} & \textbf{41.89{\scriptsize ±0.82} }& \textbf{50.01{\scriptsize ±0.85}} & \textbf{21.81{\scriptsize ±1.02}} & \textbf{32.00{\scriptsize ±0.72}} & \textbf{38.18{\scriptsize ±1.02}} & \textbf{47.93{\scriptsize ±1.26}}\\
\bottomrule
\end{tabular}
    }
    \vspace{-8pt}
    \caption{Comparison of the final average accuracy achieved
through independent inference and the use of the ensemble method on four benchmark datasets with difference memory buffer size $M$. All values are averages of 10 runs.}
    \label{tab:indaa}
    \vspace{-8pt}
\end{table*}

\begin{table*}
    \centering
    \resizebox{\textwidth}{!}{
    \begin{tabular}{lccccccccc}
\toprule
\multicolumn{1}{c}{Dataset} & \multicolumn{2}{c}{CIFAR10} & \multicolumn{3}{c}{CIFAR100} & \multicolumn{3}{c}{Tiny-ImageNet} & ImageNet-100\\
\cmidrule(lr){1-1}
\cmidrule(lr){2-3}
\cmidrule(lr){4-6}
\cmidrule(lr){7-9}
\cmidrule(lr){10-10}
\multicolumn{1}{c}{Memory Size $M$} & 500 & 1000 & 1000 & 2000 & 5000 & 2000 & 5000 & 10000 & 5000\\
\midrule
ER  & 33.16{\scriptsize ±3.50} & 20.94{\scriptsize ±6.79} & \textbf{32.65{\scriptsize ±1.78}} & \textbf{22.20{\scriptsize ±2.26}} & \textbf{13.29{\scriptsize ±1.98}} & \textbf{58.38{\scriptsize ±1.69}} & \textbf{46.87{\scriptsize ±1.60}} & \textbf{40.77{\scriptsize ±2.45}} & \textbf{23.38{\scriptsize ±2.10}} \\
ER + Ours & \textbf{30.22{\scriptsize ±3.75}} & \textbf{19.85{\scriptsize ±2.55}} & 43.28{\scriptsize ±1.67} & 29.35{\scriptsize ±1.50} & 16.88{\scriptsize ±1.99} & 69.56{\scriptsize ±1.54} & 53.13{\scriptsize ±0.85} & 42.63{\scriptsize ±0.80} & 28.48{\scriptsize ±1.50} \\
\midrule
DER++ & \textbf{24.21{\scriptsize ±2.75}} & \textbf{18.42{\scriptsize ±1.84}} & \textbf{34.49{\scriptsize ±4.39}} & \textbf{25.55{\scriptsize ±3.26}} & 20.01{\scriptsize ±2.88} & \textbf{62.03{\scriptsize ±2.83}} & \textbf{51.57{\scriptsize ±4.60}} & \textbf{49.51{\scriptsize ±3.04}} & 28.77{\scriptsize ±4.10} \\
DER++ + Ours & 25.08{\scriptsize ±2.88} & 18.47{\scriptsize ±3.12} & 42.76{\scriptsize ±1.31} & 31.13{\scriptsize ±2.41} & \textbf{18.45{\scriptsize ±2.89}} & 72.59{\scriptsize ±1.29} & 57.71{\scriptsize ±1.80} & 50.31{\scriptsize ±2.34} & \textbf{27.22{\scriptsize ±2.17}} \\
\midrule
ER-ACE & \textbf{12.72{\scriptsize ±3.56}} & \textbf{10.66{\scriptsize ±2.48}} & \textbf{12.67{\scriptsize ±1.62}} & \textbf{9.11{\scriptsize ±0.78}} & \textbf{5.92{\scriptsize ±1.09}} & \textbf{19.12{\scriptsize ±0.63}} & \textbf{17.14{\scriptsize ±0.67}} & 15.59{\scriptsize ±1.24} & \textbf{14.11{\scriptsize ±1.19}} \\
ER-ACE + Ours & 22.86{\scriptsize ±2.23} & 16.07{\scriptsize ±2.38} & 35.85{\scriptsize ±1.12} & 25.84{\scriptsize ±1.96} & 14.21{\scriptsize ±0.85} & 24.10{\scriptsize ±2.00} & 19.14{\scriptsize ±1.91} & \textbf{15.14{\scriptsize ±1.60}} & 26.02{\scriptsize ±2.33} \\
\midrule
OCM & 13.68{\scriptsize ±4.25} & 11.63{\scriptsize ±2.62} & \textbf{14.99{\scriptsize ±1.55}} & \textbf{9.16{\scriptsize ±1.75}} & \textbf{3.76{\scriptsize ±1.16}} & \textbf{26.12{\scriptsize ±1.63}} & \textbf{19.74{\scriptsize ±1.30}} & 15.92{\scriptsize ±1.47} & \textbf{3.25{\scriptsize ±0.90}} \\
OCM + Ours & \textbf{11.59{\scriptsize ±2.24}} & \textbf{9.18{\scriptsize ±2.03}} & 16.69{\scriptsize ±2.36} & 10.07{\scriptsize ±1.37} & 3.99{\scriptsize ±0.78} & 26.16{\scriptsize ±1.90} & 19.99{\scriptsize ±1.96} & \textbf{15.56{\scriptsize ±1.06}} & 8.91{\scriptsize ±1.18} \\
\midrule
GSA & \textbf{25.45{\scriptsize ±2.86}} & \textbf{16.42{\scriptsize ±3.59}} & \textbf{33.97{\scriptsize ±2.55}} & \textbf{22.74{\scriptsize ±1.83}} & \textbf{12.31{\scriptsize ±2.35}} & \textbf{27.23{\scriptsize ±2.01}} & \textbf{23.61{\scriptsize ±2.26}} & \textbf{20.58{\scriptsize ±2.09}} & \textbf{24.53{\scriptsize ±1.59}} \\
GSA + Ours & 28.47{\scriptsize ±3.08} & 19.00{\scriptsize ±2.08} & 42.41{\scriptsize ±1.44} & 31.09{\scriptsize ±1.86} & 16.77{\scriptsize ±1.87} & 64.86{\scriptsize ±1.19} & 48.23{\scriptsize ±1.28} & 35.79{\scriptsize ±1.46} & 32.77{\scriptsize ±2.07} \\
\midrule
OnPro & \textbf{17.94{\scriptsize ±3.69}} & \textbf{14.20{\scriptsize ±2.60}} & \textbf{16.76{\scriptsize ±2.47}} & \textbf{12.42{\scriptsize ±1.39}} & \textbf{6.72{\scriptsize ±0.94}} & \textbf{28.01{\scriptsize ±1.59}} & 23.52{\scriptsize ±1.75} & 20.32{\scriptsize ±1.70} & \textbf{7.59{\scriptsize ±1.17}} \\
OnPro + Ours & 19.89{\scriptsize ±4.01} & 14.62{\scriptsize ±2.75} & 28.93{\scriptsize ±2.19} & 20.23{\scriptsize ±1.03} & 10.55{\scriptsize ±1.89} & 28.21{\scriptsize ±1.58} & \textbf{20.86{\scriptsize ±1.13}} & \textbf{16.17{\scriptsize ±0.63}} & 9.90{\scriptsize ±1.93} \\
\bottomrule
\end{tabular}
    }
    \vspace{-8pt}
    \caption{Forgetting Measure (\%, lower is better) on four benchmark datasets with difference memory buffer size $M$, with and without our proposed CCL-DC scheme. The result of our method is given by the ensemble of two peer models. All values are averages of 10 runs.}
    \label{tab:fm}
    \vspace{-8pt}
\end{table*}

\paragraph{Performance with NCM classifier.}
\begin{table}
    \centering
    \resizebox{0.75\linewidth}{!}{
    \begin{tabular}{lcc}
\toprule
\multicolumn{1}{c}{Method} & NCM Acc. ↑ & Logit Acc. ↑\\
\midrule
ER & 36.56{\scriptsize ±0.60}& 31.89{\scriptsize ±1.45}\\
ER + Ours & 44.76{\scriptsize ±0.55} & 44.45{\scriptsize ±1.04} \\
\midrule
ER-ACE & 34.91{\scriptsize ±1.02} & 34.21{\scriptsize ±1.53}\\
ER-ACE + Ours & 45.62{\scriptsize ±1.04} & 45.14{\scriptsize ±1.00}\\
\midrule
OnPro & 34.32{\scriptsize ±0.95}& 33.33{\scriptsize ±0.93}\\
OnPro + Ours & 42.82{\scriptsize ±0.67} & 41.89{\scriptsize ±0.82}\\
\bottomrule
\end{tabular}
    }
    \caption{Final average accuracy on CIFAR-100 ($M=$2k), with and without NCM classifier.}
    \vspace{-8pt}
    \label{tab:ncm}
\end{table}

Besides t-SNE, we can evaluate the feature discrimination using the clustering methods. We remove the final FC classifier and use Nearest-Class-Mean (NCM) classifier with intermediate representations. Table~\ref{tab:ncm} demonstrates that CCL-DC can greatly enhance the NCM accuracy, which evidences the capability of CCL-DC in improving feature discrimination.

\begin{figure}[t]
  \centering
   \vspace{-8pt}
   \includegraphics[width=0.8\linewidth]{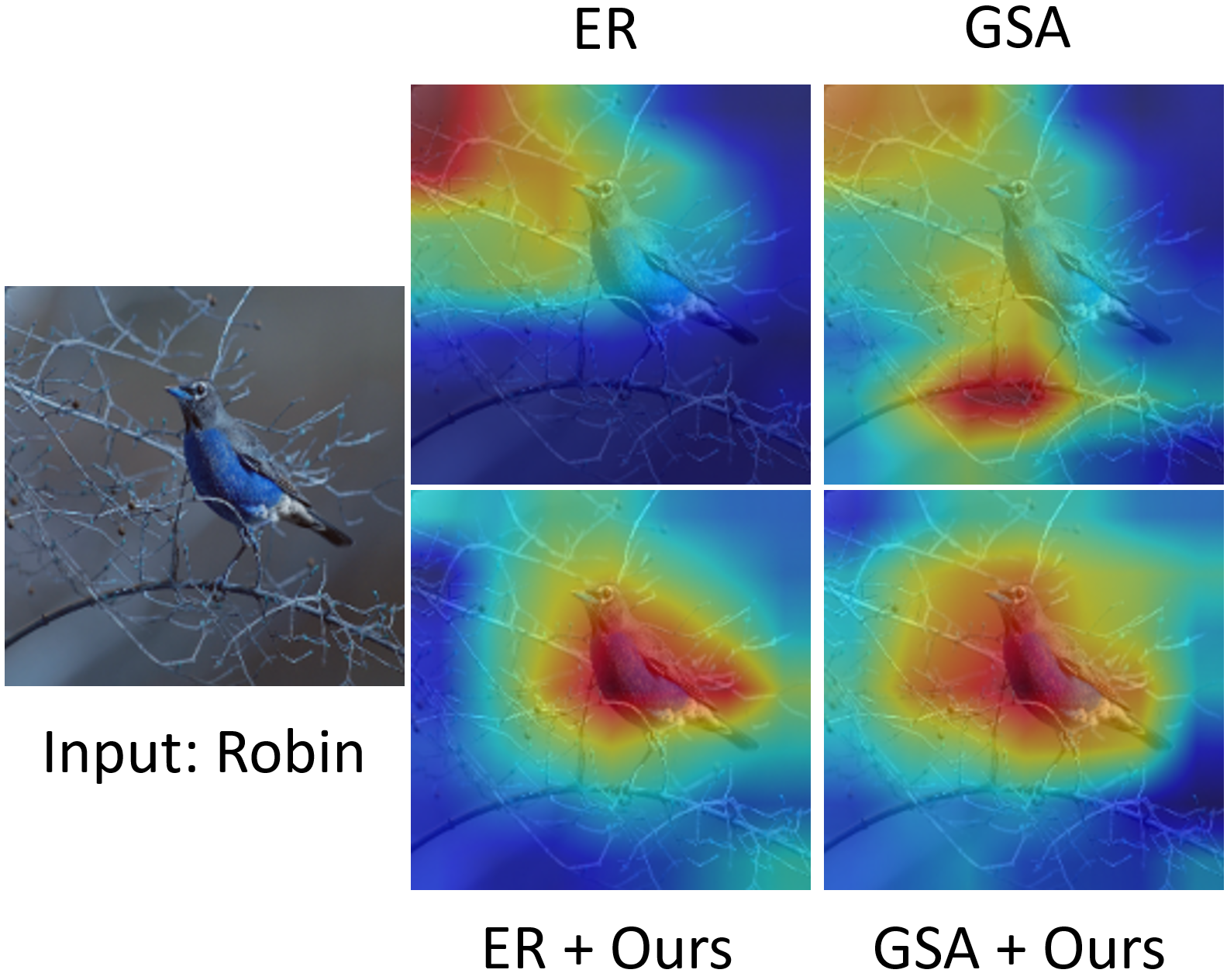}
   \caption{GradCAM++ visualization on the training set of ImageNet-100 ($M=$5k). Shortcut learning exists in the baseline methods despite making correct predictions.}
   \label{fig:CAM}
   \vspace{-8pt}
\end{figure}

\paragraph{GradCAM++ visualization.}
Shortcut learning is another commonly observed issue that hinders the generalization capability of continual learners~\cite{wei2023online}. In Fig.~\ref{fig:CAM}, we use GradCAM++ on the training set of ImageNet-100 ($M=$5k) at the end of the training of ER and GSA. Although both ER and GSA make correct predictions, we observed that they focus on irrelevant objects, which indicates a tendency toward shortcut learning. Also, we can see that by integrating CCL-DC, the shortcut learning can be greatly alleviated. 

\section{Counterintuitive performance of ER-ACE}

As shown in Table~\ref{tab:la}, ER-ACE suffers from counterintuitive performance on plasticity, especially when the task number is large (\textit{e.g.}, TinyImageNet experiments). This is because ER-ACE employs Asymmetric Cross-Entropy loss (ACE) during the training of batch images. ACE manually masks out the old classes for batch image training, which reduces the feature drift of old classes and enhances ER-ACE's stability, as stated in the original paper. However, ACE cuts the gradient for old classes in classification loss, which limits the optimizer's maneuverability in the final classification layer. This loss of maneuverability is significant when there are many tasks involved, and thus we may observe the LA close to 0 in the later stages of training. Despite the low plasticity, ER-ACE has a good overall performance because: (1) Memory replay partially compensates for this loss in terms of plasticity (Learner can still learn from samples in the memory buffer), and (2) ER-ACE has higher stability.
Additionally, we witness a major stability drop in ER-ACE when incorporating CCL-DC. As indicated in the Algorithm~\ref{code:pseudo_code}, although we also use ACE in the classification loss of DC, we do not perform masking in the distillation loss. The distillation between probability distributions of peer models retrieves some plasticity, but it also leads to extra feature drift, which hurts the stability to some extent.

\section{Experiment Details}
\label{sec:detail}

\paragraph{Dataset}
As mentioned in Sec.~\ref{sec:experiments}, we use four datasets to evaluate the effectiveness of our method. The original datasets are split into several tasks of disjoint classes. The detailed information about dataset split and task allocation is as follows:

\noindent \textbf{CIFAR-10}~\cite{krizhevsky2009learning} has 10 classes with 50,000 training samples and 10,000 test samples. Images are sized at $32 \times 32$. In our experiments, it is split into five non-overlapping tasks with two classes per task.

\noindent \textbf{CIFAR-100}~\cite{krizhevsky2009learning} has 100 classes with 50,000 training samples and 10,000 test samples. The images are $32 \times 32$ in size. It is split into 10 disjoint tasks with 10 classes per task.

\noindent \textbf{TinyImageNet}~\cite{le2015tiny} has 200 classes with 100,000 training samples and 10,000 test samples. Images are sized at $64 \times 64$. It is split into 100 non-overlapping tasks with two classes per task. 

\noindent \textbf{ImageNet-100}~\cite{hou2019learning} is the subset of ImageNet-1k~\cite{deng2009imagenet} containing 100 classes. We follow~\cite{imagenet100pytorch} for the class selection. The images are $224 \times 224$ in size. It is split into 10 disjoint tasks with 10 classes per task.

\paragraph{Task Sequence.} In online CL, some work uses a fixed task sequence throughout all runs to evaluate the performance, for the sake of fair comparison. However, we found that the evaluation heavily depends on the task order. For fair comparison, we randomize the allocation of classes to tasks and the sequence of tasks using 10 fixed random seeds (for 10 runs in our experiments). This ensures our evaluation result is not biased to task difficulty.

\paragraph{Data augmentation for baseline methods.}
Data augmentation has been demonstrated to be successful in improving the performance of online CL~\cite{rolnick2019experience, buzzega2020dark, guo2022online}. However, methods benefit differently from different augmentation intensities, and some methods may gain more performance with simple augmentations instead of complicated ones. Thus, to achieve optimal performance for comparison, we involve two different augmentation strategies for baseline methods:

\noindent \textbf{1. Partial augmentation strategy.} The partial augmentation is a strategy with weak augmentation. It comprises random cropping with $p=0.5$, followed by random horizontal flip with $p=0.5$.

\noindent \textbf{2. Full augmentation strategy.} The full strategy is a superset of the partial strategy. It consists of random cropping, horizontal flipping, color jitter, and random grayscale. The parameter of color jitter is set to (0.4, 0.4, 0.4, 0.1) with $p=0.8$, while the probability of random grayscale is 0.2.

For fair comparison, the models trained with CCL-DC also employ the same data augmentation strategy in the baseline loss part, as illustrated in Algorithm~\ref{code:pseudo_code}.

\paragraph{Hyperparameter search for baselines.}
For hyperparameters in the baseline methods, as indicated in Sec.~\ref{sec:experiments}, we perform a hyperparameter search on CIFAR-100 ($M=$2k) for the baseline methods. Table~\ref{tab:hpsearch} shows the exhaustive list of the grid search. Note that we used the hyperparameters from the original OCM paper to reduce the hyperparameter search space due to computational constraints. For fair comparison, after finding the optimal hyperparameters for the baseline methods, we apply the same hyperparameters when incorporating CCL-DC. 

\begin{figure}
\begin{center}
    \subfloat[Performance when $M=15$]{
       \includegraphics[width=0.48\linewidth]{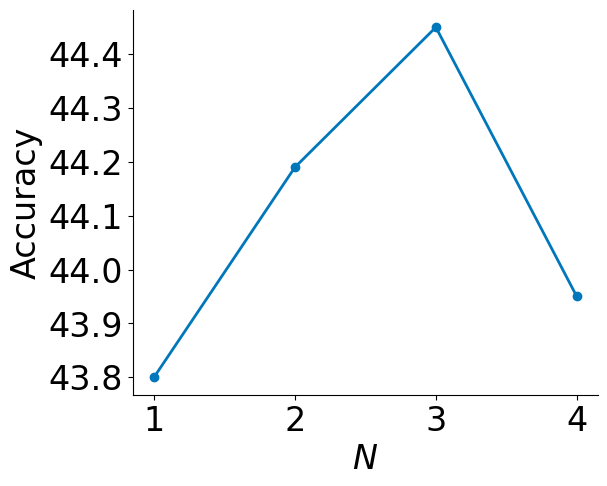}}
    \label{1a}\hfill
    \subfloat[Performance when $N=3$]{
        \includegraphics[width=0.48\linewidth]{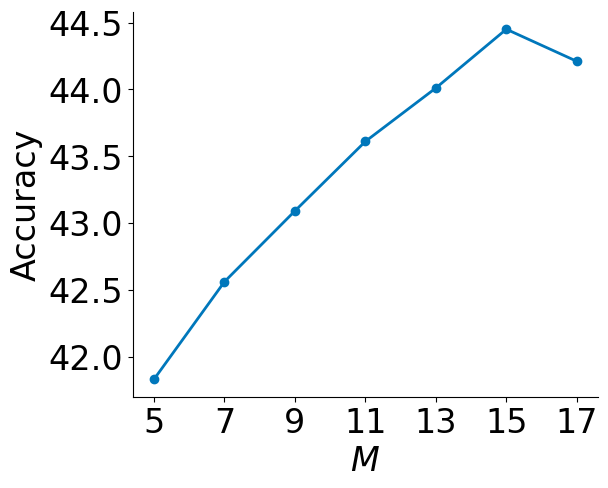}}
    \label{1b}
    \caption{The impact of $N$ and $M$ in RandAugment on the performance for ER + \textit{Ours} on CIFAR-100 ($M=$2k). As shown in the figure, the best performance is achieved with $N = 3$ and $M = 15$. All numbers are averaged over 10 runs.}
    \label{fig:randaug} 
    \vspace{-18pt}
\end{center}
\end{figure}

\begin{figure}
\begin{center}
    \subfloat[Performance when $\lambda_2=2$]{
       \includegraphics[width=0.48\linewidth]{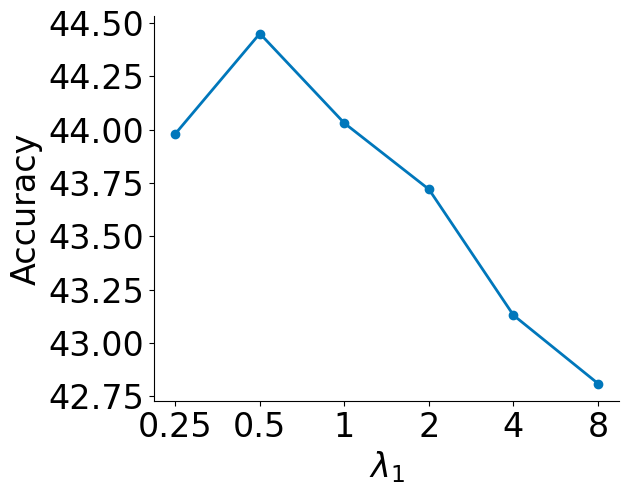}}
    \label{1a}\hfill
    \subfloat[Performance when $\lambda_1=0.5$]{
        \includegraphics[width=0.48\linewidth]{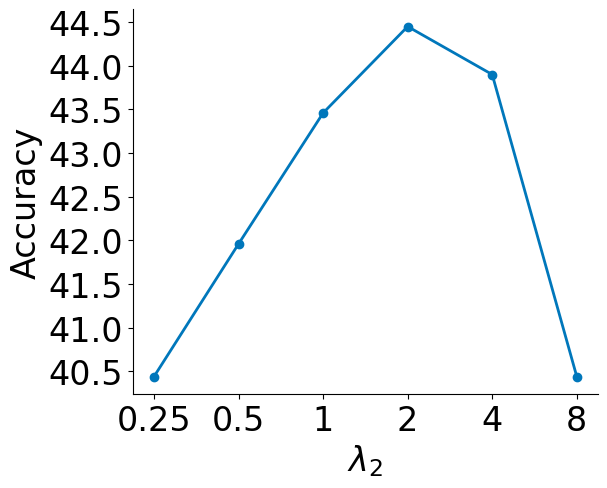}}
    \label{1b}
    \caption{The impact of $\lambda_1$ and $\lambda_2$ on the performance for ER + \textit{Ours} on CIFAR-100 ($M=$2k). As shown in the figure, the best performance is achieved with $\lambda_1=0.5$ and $\lambda_2=2$. All numbers are averaged over 10 runs.}
    \label{fig:lam} 
    \vspace{-18pt}
\end{center}
\end{figure}

\paragraph{Hyperparameter search for CCL-DC.} CCL-DC also has four unique hyperparameters, including $N$, $M$ in RandAugment of DC, and $\lambda_1$, $\lambda_2$ in Eq.~\ref{eq:CCL} and Eq.~\ref{eq:DC}.

In CCL-DC, we use RandAugment to generate samples with different difficulty. Thus, two additional hyperparameters in RandAugment ($N$ and $M$) are involved in CCL-DC. Since the transformation intensity ($N$ and $M$) is highly related to the dataset instead of the baseline method, We conduct a hyperparameter search for every dataset with ER + CCL-DC and apply the same hyperparameter across all baseline methods when incorporating CCL-DC. We searched in 4 settings of $N$ and 7 settings of $M$ (\textit{i.e.}, $N=\{1,2,3,4\}$ and $M=\{5,7,9,11,13,15,17\}$). With our grid search, we find that ($N=3, M=15$) is the best for CIFAR-10 and CIFAR-100. ($N=1, M=11$) achieves the best results on Tiny-ImageNet and ($N=3, M=11$) is the best for ImageNet-100. 
We also visualize some of the experimental results on CIFAR-100, as shown in Fig.~\ref{fig:randaug}.

CCL-DC also comprises two hyperparameters $\lambda_1$ and $\lambda_2$ in Eq.~\ref{eq:CCL} and Eq.~\ref{eq:DC}. Similar to the hyperparameter search strategy we do for baseline hyperparameters, for each baseline method with CCL-DC, we initiate another hyperparameter search for $\lambda_1$ and $\lambda_2$ on CIFAR-100 ($M=$2k) and apply the hyperparameter to all of the settings. We searched from $\lambda_1,\lambda_2 = \{0.25, 0.5, 1, 2, 4, 8\}$. We visualize some experimental results in Fig~\ref{fig:lam}.

\begin{table*}
    \centering
    \resizebox{0.6\linewidth}{!}{
    \begin{tabular}{ccc}
\toprule
Method & HP & Values \\
\midrule
\multirow{5}{*}{ER}& optimizer & [SGD, AdamW]\\
    & lr & [0.1, 0.05, 0.01, 0.005, 0.001, 0.0005, 0.0001] \\
    & weight decay & [0, 1e-4] \\
    & momentum (for SGD) & [0, 0.9] \\
    & aug. strat. & [partial, full] \\
\midrule
\multirow{7}{*}{DER++}& optimizer & [SGD, AdamW]\\
    & lr & [0.1, 0.05, 0.01, 0.005, 0.001, 0.0005, 0.0001] \\
    & weight decay & [0, 1e-4] \\
    & momentum (for SGD) & [0, 0.9] \\
    & aug. strat. & [partial, full] \\
    & alpha & [0.1, 0.2, 0.5, 1.0] \\
    & beta & [0.5, 1.0] \\
\midrule
\multirow{5}{*}{ER-ACE}& optimizer & [SGD, AdamW]\\
    & lr & [0.1, 0.05, 0.01, 0.005, 0.001, 0.0005, 0.0001] \\
    & weight decay & [0, 1e-4] \\
    & momentum (for SGD) & [0, 0.9] \\
    & aug. strat. & [partial, full] \\
\midrule
\multirow{4}{*}{OCM}& optimizer & [AdamW]\\
    & lr & [0.001] \\
    & weight decay & [1e-4] \\
    & aug. strat. & [partial, full] \\
\midrule
\multirow{5}{*}{GSA}& optimizer & [SGD, AdamW]\\
    & lr & [0.1, 0.05, 0.01, 0.005, 0.001, 0.0005, 0.0001] \\
    & weight decay & [0, 1e-4] \\
    & momentum (for SGD) & [0, 0.9] \\
    & aug. strat. & [partial, full] \\
\midrule
\multirow{5}{*}{OnPro}& optimizer & [SGD, AdamW]\\
    & lr & [0.1, 0.05, 0.01, 0.005, 0.001, 0.0005, 0.0001] \\
    & weight decay & [0, 1e-4] \\
    & momentum (for SGD) & [0, 0.9] \\
    & aug. strat. & [partial, full] \\
\bottomrule
\end{tabular}
    }
    \caption{Exhaustive list of hyperparameters searched on CIFAR-100 ($M=$2k).}
    \label{tab:hpsearch}
\end{table*}

\begin{figure*}[t]
  \centering
   \includegraphics[width=0.7\linewidth]{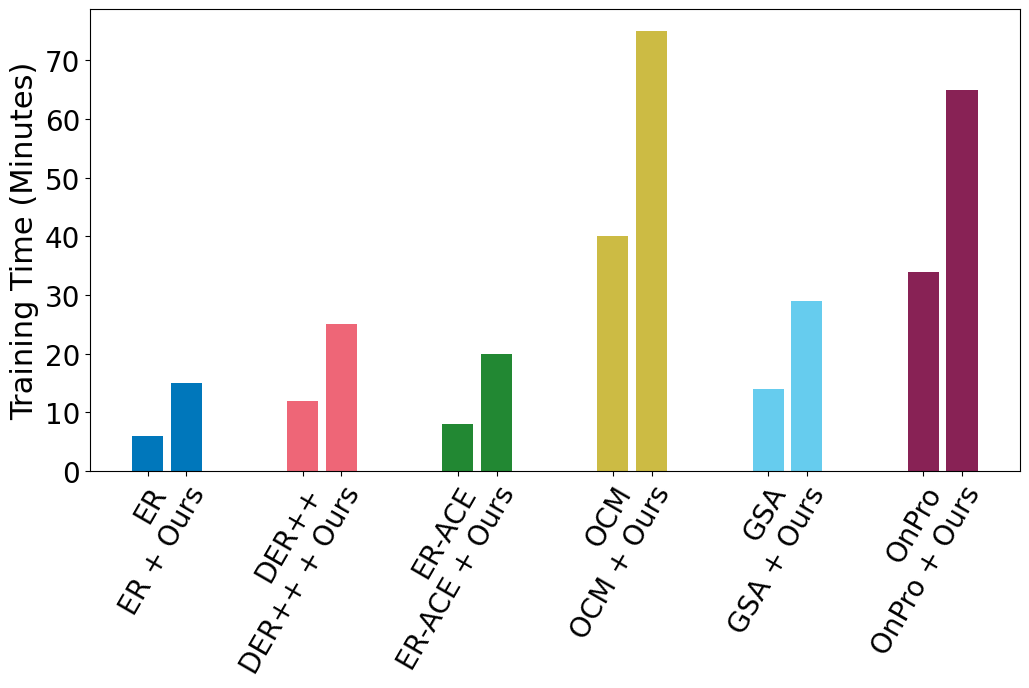}
   \caption{Training time of each method on CIFAR-100 ($M=$2k).}
   \label{fig:computation}
\end{figure*}

\paragraph{Hardware and Computation.}
All the experiments in our work are conducted on NVIDIA A100 GPUs. Fig.~\ref{fig:computation} shows the training time for each method with and without CCL-DC on CIFAR-100 ($M=$2k).

\end{document}